\documentclass[10pt,twocolumn,letterpaper]{article}

\usepackage{cvpr}
\usepackage{times}
\usepackage{epsfig}
\usepackage{graphicx}
\usepackage{amsmath}
\usepackage{amssymb}
\usepackage{verbatim}

\usepackage{booktabs,tabulary,overpic}
\usepackage{multirow}
\usepackage{stfloats}
\usepackage{tabularx}
\usepackage{algorithm}
\usepackage{algpseudocode}
\usepackage{subcaption}
\usepackage{bm}
\usepackage{mathtools}
\usepackage[usenames, dvipsnames]{color}
\usepackage{numdef}
\usepackage{soul}
\usepackage{adjustbox}
\usepackage{enumitem}
\setlist{nosep}

\usepackage[T1]{fontenc}
\usepackage[utf8]{inputenc}
\usepackage[font=small,labelfont=bf]{caption}

\definecolor{whitesmoke}{rgb}{0.96, 0.96, 0.96}
\definecolor{antiquewhite}{rgb}{0.98, 0.92, 0.84}
\definecolor{citecolor}{RGB}{34,139,34}

\usepackage[pagebackref=true,breaklinks=true,letterpaper=true,colorlinks,citecolor=citecolor,bookmarks=false]{hyperref}

\cvprfinalcopy % *** Uncomment this line for the final submission

\def\cvprPaperID{2966} % *** Enter the CVPR Paper ID here

\begin{document}
%%%%%%%%%%%%%%%%my commands%%%%%%%%%%%%%%%%%%%%%%%%%%%

\makeatletter\newcommand\footnoteref[1]{\protected@xdef\@thefnmark{\ref{#1}}\@footnotemark}\makeatother
\newcommand{\app}{\raise.17ex\hbox{$\scriptstyle\sim$}}                         
\newcommand{\ncdot}{{\mkern 0mu\cdot\mkern 0mu}}                                
\def\x{\times}                                                                  
\newcolumntype{x}[1]{>{\centering\arraybackslash}p{#1pt}}                       
\newcommand{\dt}[1]{\fontsize{8pt}{.1em}\selectfont \emph{#1}}                  
\newlength\savewidth\newcommand\shline{\noalign{\global\savewidth\arrayrulewidth
  \global\arrayrulewidth 1pt}\hline\noalign{\global\arrayrulewidth\savewidth}}  
\newcommand{\tablestyle}[2]{\setlength{\tabcolsep}{#1}\renewcommand{\arraystretch}{#2}\centering\footnotesize}
\makeatletter\renewcommand\paragraph{\@startsection{paragraph}{4}{\z@}          
  {.5em \@plus1ex \@minus.2ex}{-.5em}{\normalfont\normalsize\bfseries}}\makeatother

\newcommand{\figthree}[3]{
  \subfigure{  
  \begin{minipage}{5cm}
    \centering   
    \includegraphics[scale=0.4]{#1}            
  \end{minipage}
  }
   \quad \quad
  \subfigure{  
    \begin{minipage}{5cm}
      \centering    
      \includegraphics[scale=0.4]{#2}                
     \end{minipage}
   }
   \quad \quad
   \subfigure{
    \begin{minipage}{5cm}
     \centering    \includegraphics[scale=0.4]{#3}    
    \end{minipage}
   }
}
\newcommand{\figtwo}[2]{
  \subfigure{  
  \begin{minipage}{8cm}
    \centering   
    \includegraphics[scale=0.6]{#1}            
  \end{minipage}
  }
   \quad \quad
  \subfigure{  
    \begin{minipage}{8cm}
      \centering    
      \includegraphics[scale=0.6]{#2}                
     \end{minipage}
   }
}
\newcommand{\figref}[1]{Figure~\ref{#1}}
\newcommand{\tableref}[1]{Table~\ref{#1}}
\newcommand{\equationref}[1]{Equation~\ref{#1}}
\newcommand{\etalcite}[1]{~\etal~\cite{#1}}
\newcommand{\sota}{state-of-the-art}

% phrases
\newcommand{\handdataset}{InterHand}
\newcommand{\humandataset}{Human3.6M~\cite{ionescu2013human3}}
\newcommand{\dtu}{DTU}
\newcommand{\ours}{epipolar transformer}
\newcommand{\Ours}{Epipolar transformer}
\newcommand{\OURS}{Epipolar Transformer}
\newcommand{\mainview}{reference view}
\newcommand{\otherview}{source view}
\newcommand{\mainnotation}{\textrm{reference}}
\newcommand{\othernotation}{\textrm{source}}
\newcommand{\mainnotationabbr}{\textrm{ref}}
\newcommand{\othernotationabbr}{\textrm{src}}
\newcommand{\crossview}{cross-view~\cite{qiu2019cross}}
\newcommand{\crossviewfusion}{cross-view fusion~\cite{qiu2019cross}}
\newcommand{\rpsm}{RPSM~\cite{qiu2019cross}}
\newcommand{\augmentation}{$^+$}

%numbers
\newcommand{\handdatasetrgbviews}{34}
\num\def\handbaseline36view{5.46}
\num\def\handour36view{4.91}
\newcommand{\humansmallresnetfiftydltjdr}{97.01}
\newcommand{\humansmallresnetfiftydltaugjdr}{98.25}
\newcommand{\humansmallresnetfiftydlt}{33.1} %33.10
\newcommand{\humansmallresnetfiftydltaug}{30.4} %30.43
\newcommand{\humansmallresnetfiftyrpsm}{26.9} %26.94
\newcommand{\humanextrabigresnetonefifydlt}{19.0}

%math def
\newcommand{\sampler}{\mathcal{S}}
\newcommand{\samplesize}{K}
\newcommand{\sampledfeat}{Y}
\newcommand{\mainloc}{p}
\newcommand{\otherloc}{p'}
\newcommand{\mainproj}{M}
\newcommand{\otherproj}{M'}
\newcommand{\mainintrinsic}{K}
\newcommand{\otherintrinsic}{K'}
\newcommand{\maincameracenter}{C}
\newcommand{\othercameracenter}{C'}
\newcommand{\mainimage}{\mathcal{I}}
\newcommand{\otherimage}{\mathcal{I}'}
\newcommand{\otherepipolarline}{l}
\newcommand{\moduleoutput}{z}
\newcommand{\weightz}{W_z}
\newcommand{\epipolarsampler}{epipolar sampler}
\newcommand{\Epipolarsampler}{Epipolar sampler}
\newcommand{\channels}{256}

\newcommand{\yihui}[1]{{\color{blue} Yihui: #1}}
\newcommand{\rui}[1]{{\color{citecolor} Rui: #1}}
%%%%%%%%% TITLE
\title{Epipolar Transformers}
% \title{\OURS\ for Multi-view Pose Estimation}

\author{Yihui He$^*$ \quad Rui Yan\thanks{Equal contribution} \quad Katerina Fragkiadaki\\
Carnegie Mellon University\\
Pittsburgh, PA 15213\\
{\tt\small \{he2@alumni,ruiyan@alumni,katef@cs\}.cmu.edu}
\and
Shoou-I Yu\\
Facebook Reality Labs\\
Pittsburgh, PA 15213\\
{\tt\small shoou-i.yu@fb.com}
}

\maketitle
% \thispagestyle{empty}

%%%%%%%%% ABSTRACT
\begin{abstract}
A common approach to localize 3D human joints in a synchronized and calibrated multi-view setup consists of two-steps: 
(1) apply a 2D detector separately on each view to localize joints in 2D, and
(2) perform robust triangulation on 2D detections from each view to acquire the 3D joint locations.
However, in step 1, the 2D detector is limited to solving challenging cases 
which could potentially be better resolved in 3D,
such as occlusions and oblique viewing angles, 
purely in 2D without leveraging any 3D information.
%Subsequently, step 2 will perform poorly when the predictions from step 1 are noisy or not consistent in 3D.
Therefore, we propose the differentiable ``\ours'', which enables the 2D detector to leverage 3D-aware features to improve 2D pose estimation.
The intuition is: given a 2D location $\mainloc$ in the current view, 
we would like to first find its corresponding point $\otherloc$ in a neighboring view,
and then combine the features at $\otherloc$ with the features at $\mainloc$,
thus leading to a 3D-aware feature at $\mainloc$.
Inspired by stereo matching, the \ours$\,$ leverages epipolar constraints and feature matching to approximate the features at $\otherloc$.
%The key advantages of the \ours\ are: 
%(1) it has minimal learnable parameters, 
%(2) it can be easily plugged into existing networks,
%(3) it is interpretable, i.e., we can analyze the location $\otherloc$ to understand whether matching over the epipolar line was successful, and (4) it can be potentially transferred between new camera setups without additional training.
Experiments on \handdataset\ and \humandataset\ show that our approach has consistent improvements over the baselines.
%Furthermore, qualitative analysis shows that the deep features learned in this fashion can perform matching better than RGB features.
Specifically, in the condition where no external data is used, our Human3.6M model trained with ResNet-50 backbone and image size 256$\times$256 outperforms \sota\ by 4.23mm and achieves MPJPE \humansmallresnetfiftyrpsm\ mm. Code is available\footnote{\href{https://github.com/yihui-he/epipolar-transformers}{github.com/yihui-he/epipolar-transformers}}.
\end{abstract}

%%%%%%%%% BODY TEXT
\section{Introduction}
% \begin{figure}[!t]
% \begin{center}
%     \includegraphics[width=0.8\linewidth]{imgs/epipolar_sample.png}
%     \vspace{-2mm}
%     \caption{Given $p$ in the \mainview, the proposed \Ours\ leverages deep features along the epipolar line to find correspondence $p'$ in the \otherview. Then the features at $p'$ are fused with the features at $p$. Detailed visualizations are in \figref{fig:visualization}. (\textit{best view in color})}
%     \label{fig:geometry}
% \end{center}
% \vspace{-5mm}
% \end{figure}

\begin{figure}
    % \begin{center}
    \centering
    \includegraphics[width=1.05\linewidth]{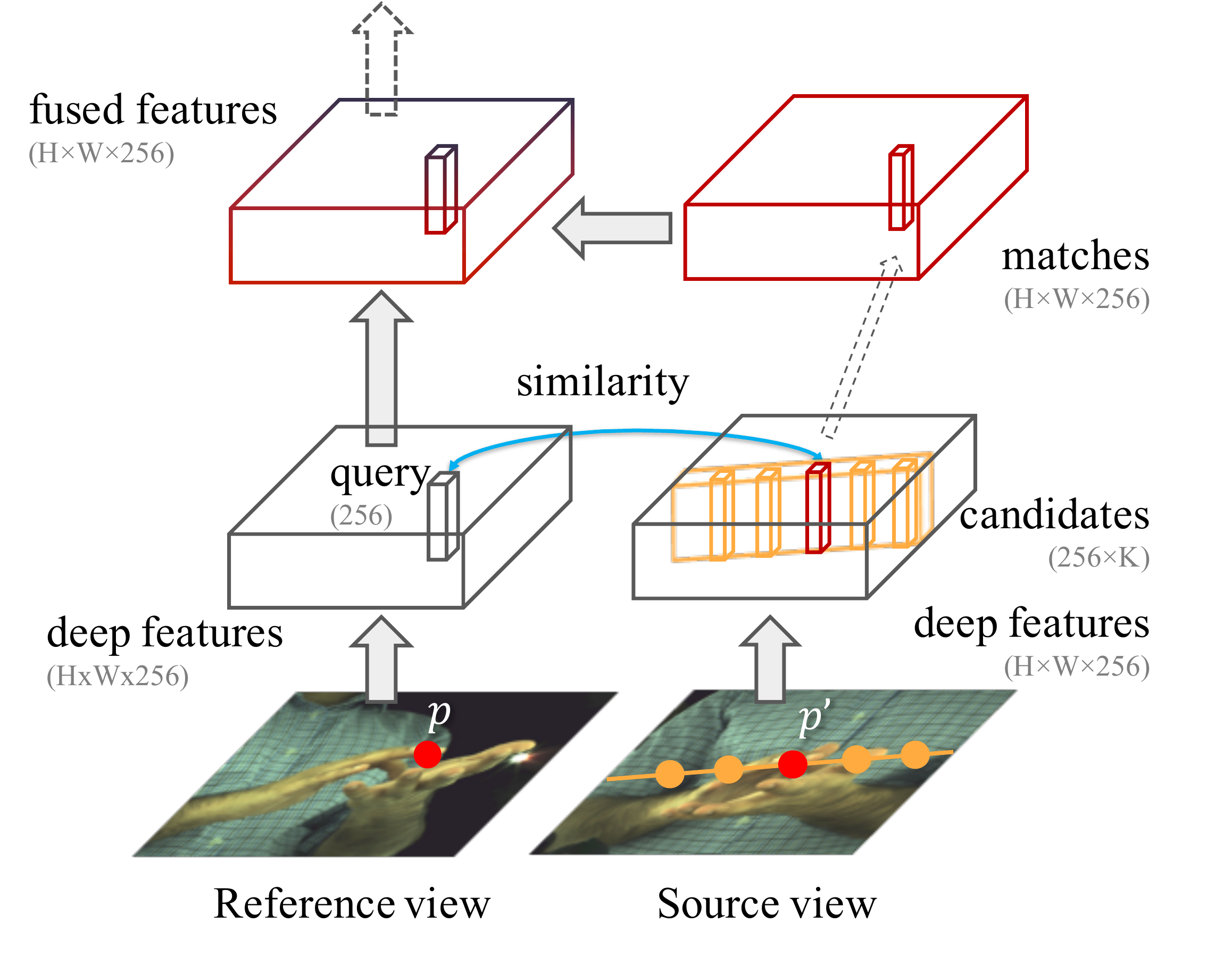}
    \vspace{-4mm}
    \caption{Overview of the proposed \ours, which enables 2D detectors to leverage 3D-aware features for more accurate pose estimation. For a query vector (\eg, with length \channels) on the intermediate deep feature maps of the \mainview\  (H$\times$W$\times$\channels), we extract \samplesize\ samples 
    %(\channels$\times$\samplesize) 
    along the corresponding epipolar line in the \otherview. Dot-product and softmax are used to compute similarity between the query and sampled vectors, which in turn is used to compute the corresponding feature. The corresponding feature is then fused with the \mainview\ feature to arrive at a 3D-aware feature for the \mainview. %(\textit{better view in color}) %avoid one extra line... 
\textbf{}    }
    \label{fig:arch}
    % \end{center}
\end{figure}

% With advances in deep neural networks, the performance of 3D hand and human pose estimators have significantly improved. 
% The 3D pose estimation task can be divided into two categories:
In order to estimate the 3D pose of a human body or hand, there are two common settings. The first setting is single-view 3D pose estimation~\cite{zimmermann2017learning,zhang2019end,boukhayma20193d,ge20193d}, where the algorithm directly estimates 3D pose from a single image. This is extremely challenging due to the ambiguity in depth when only one view is available.
% but This is an ill-posed problem due to the ambiguity in depth, which can be alleviated through multi-view pose estimation. This paper focuses on the latter.
% multi-view pose estimation utilizes multiple images from different cameras taken at the same time. This work focuses on the latter. 
% The 3D pose estimation task is to localize keypoints on a target object
The second setting, which is the focus of our paper, is multi-view 3D pose estimation, where the algorithm can leverage
multiple synchronized and geometrically calibrated views to resolve depth ambiguity.
A common framework~\cite{simon2017hand,jafarian2018monet} to resolve depth ambiguity and accurately estimate the 3D position of joints 
follows a two-step process:
(1) apply a 2D pose detector on each view separately to localize joints in 2D, 
and (2) perform robust triangulation based on camera calibration and 2D detections from each view to acquire the 3D position of joints.
Robust triangulation is required as the prediction of the 2D pose detector could be incorrect or missing due to occlusions.
One main disadvantage of this framework is that the detection in step~1 predicts keypoint positions \textit{independently from all other views}. Thus, challenging cases 
that could potentially be better resolved in 3D, 
such as occlusions and viewing the scene from oblique angles, are all resolved by the detector in 2D \textit{without utilizing any 3D information}. 
This could lead to inaccurate detections that are inconsistent in 3D, or the network might require more capacity and training data to resolve these challenging cases.
%estimating 2D poses from each single view image individually and merging the 2D pose information to calculate the final 3D pose (\eg, triangulation). The quality of 3D pose estimation greatly relies on the quality of the estimated 2D poses, which is often difficult in practice due to occlusion or motion blur.

%Therefore, in this paper, we explore the possibility of leveraging 3D information not only
%on the final 2D detections, but also on the 
%The intuition of our proposed approach is 
%shown in \figref{fig:arch}: given a 2D location $p$ in the \mainview, 
%we first find its corresponding point $p'$ in the \otherview,
%and then combine the features at $p'$ with the features at $p$.
%In this way, the intermediate features of a 2D detector can take advantage of features from neighboring views to compute

To this end, we propose the fully differentiable ``\ours'' module, which enables a 2D detector to gain access to 3D information in the \textit{intermediate layers} of the 2D detector itself, and not only during the final robust triangulation phase.
The \ours\ augments
the intermediate features of a 2D detector for a given view (\mainview) with features from neighboring views (\otherview), thus making the intermediate features 3D-aware as shown in \figref{fig:arch}.
%which is a fully differentiable module that takes a feature at $p$ in the \mainview, and fuses it with the estimated feature at $p'$ in the \otherview to augment the feature with 3D information.
%Inspired by stereo matching, 
%The intuition of our proposed approach is shown in \figref{fig:arch}.
To compute the 3D-aware intermediate feature at location $p$ in the \mainview, 
we first find the point corresponding to $p$ in the \otherview: $p'$, and then fuse the feature at $p'$
with the feature at $p$ to get a 3D-aware feature.
However, we do not know where the correct $p'$ is, so in order to approximate the feature at $p'$,
we first leverage the epipolar line generated by $p$ in the \otherview\ to limit the potential locations of $p'$.
Then, we compute the similarity between the feature at $p$ and the features sampled along the epipolar line.
Finally, we perform a weighted sum over the features along the epipolar line
as an approximation of the feature at $p'$.
The weights used for the weighted sum are the feature similarity.
In order to fuse the features at $p$ and $p'$,
we propose multiple methods inspired by non-local networks \cite{wang2018non}.
%to fuse the two features and arrive at the 3D-aware intermediate features for location $p$.
Note that the aforementioned operation is done densely for all locations in an intermediate feature map,
so the final output of our module is a set of 3D-aware intermediate features that have the same dimensions as the input feature map.

Since the \ours\ is fully differentiable and outputs features with the same dimensions as the input, 
it can be flexibly inserted into desired locations of a 2D pose detection network and trained end-to-end.
The inputs to our network are geometrically calibrated and synchronized multi-view images, and the output of the network is
2D joint locations, which can then further be triangulated to compute 3D joint locations.
%Note that since the \ours\ only operates on the intermediate features of the network, 
%the final output of the detector is the 2D location of joints and not 3D.
Note that even though our network outputs 2D joint locations, our network has access to both 2D and 3D features, thus empowering it to leverage more information to achieve more accurate 2D predictions.

%\todo{The last sentence not clear enough?}

%The key advantages of the \ours$\,$ is: 
%(1) it is differentiable,
%(2) it has minimal learnable parameters, 
%(3) it can be easily plugged into existing networks,
%and (4) it is easily interpretable, i.e., we can analyze the location $p'$ to understand whether matching over the epipolar line was successful.

To evaluate the performance of our \ours, we have conducted experiments on \humandataset\ and \handdataset. On \humandataset, we achieve \humansmallresnetfiftyrpsm\ mm mean per-joint position error when using the ResNet-50 backbone on input images of resolution 256$\times$256 and trained without external data. This outperforms the \sota, Qiu~\etal~\cite{qiu2019cross} from ICCV'19 by 4.23 mm.
\handdataset$\,$ is an internal multi-view hand dataset where we also consistently outperform the baselines. %we consistently achieve around 10\% relative improvement compared to Qiu~\etal~\cite{qiu2019cross} and the baseline detector using Resnet-50~\cite{resnet},
% using the Hourglass~\cite{newell2016stacked},
%which does not utilize neighboring view information.
%We also performed qualitative analysis which showed that the deep features we learn through the \ours, promotes features that are more effective in matching than RGB features, as shown in \figref{fig:visualization}.

%The proposed \ours$\,$ has multiple advantages, including:
In sum, the strengths of our method are as follows.
\begin{enumerate}
    \item The \ours\ can easily be added into existing network architectures given it is fully differentiable and the output feature dimensions are the same as the input.
    \item The \ours\ contains minimal learnable parameters (parameter size is $C$-by-$C$, where $C$ is input feature channel size).
    \item The \ours\ is interpretable because one can analyze the feature similarity along the epipolar line to gauge whether the matching is successful.
    \item The network learned with the \ours\ could generalize to new multi-camera setups that are not included in the training data as long as if the intrinsics and extrinsics are provided.
\end{enumerate}
%The \ours\ not only enables features of 2D detectors to be influenced by features from other views, 
%but also potentially promotes features to be coherent across views.

%In summary, our contributions are as follows:
%\begin{enumerate}
%    \item We propose the \ours, which is a differentiable module that enables existing 2D pose detectors to gain access to 3D-aware features, thus leading to more accurate 3D pose predictions. 
%    \item Experiments show that our proposed method improves upon the \sota\ on both hand and human pose estimation tasks.
%    \item We perform detailed ablation studies to analyze the \ours and understand the effect of different design choices.
%\end{enumerate}

%Experiments on \handdataset\ and \humandataset\ show that our approach has consistent improvements over the baselines. Specifically, our ResNet-50 with image size 256$\times$256 outperforms \sota\ by a large margin and achieves MPJPE \textbf{\humansmallresnetfiftyrpsm mm} on \humandataset. 

%In this work, we propose \OURS\ to estimate 2D poses using information from neighboring views. From epipolar geometry (illustrated in \figref{fig:geometry}), we know that corresponding features can be found on the epipolar line of neighboring view. \Ours\ fuses features that are similar to the features on the \mainview. This module can be plugged into the middle of existing architectures.

%%%%%%%%%%%%%%%%%%%%%%%%%%%%%%%%%%%%%%%%%%%%%%%%%%%%%%%% Related work %%%%%%%%%%%%%%%%%%%%%%%%%%%%%%%%%%%%%%%%%%%%%%%%%%%%%%%
\section{Related Work}
\paragraph{Multi-view 3D Human Pose Estimation:}
There are many methods proposed for multi-view human pose estimation. 
Pavllo\etalcite{pavllo20193d} proposed estimating 3D human pose in video via dilated temporal convolutions over 2D keypoints. Rhodin\etalcite{rhodin2018learning} proposed to leverage multi-view constraints as weak supervision to enhance a monocular 3D human pose detector when labeled data is limited.
Our method is most similar to Qiu~\etal~\cite{qiu2019cross} and
Iskakov~\etal~\cite{iskakov2019learnable}, thus we provide a more detailed comparison in the following paragraphs.

 Qiu~\etal~\cite{qiu2019cross}
 proposed to fuse features from other views through learning a fixed attention weight for all pairs of pixels for each pair of views. The advantage of this method is that camera calibration is no longer required. However, the disadvantages are (1) more data from each view is needed to train this attention weight, (2) there are significantly more weights to learn when the number of views and the image resolution increases, and (3) during test time, if the multi-camera setup changes, then the attention learned during training time is no longer applicable. On the other hand, although the proposed \ours\ relies on camera calibration, it only adds minimal learnable parameters. This makes it significantly easier to train, and thus has less demand on the number of training images per view (\tableref{tab:hand_fusion}). Furthermore, the network trained with the \ours\ can be applied to an unseen multi-camera setup without additional training as long as if the calibration parameters are provided.

%which is in the same vein as our work. The key difference is Qiu~\etal~\cite{qiu2019cross} 
%Note that Qiu~\etal~\cite{qiu2019cross} also performs cross-view deep feature fusion, but epipolar geometry was not used to compute the cross-view fixed attention weights.
%For multi-view keypoint estimation, Qiu~\etal~\cite{qiu2019cross} proposed to fuse all feature pixels from other views with fixed attention for each camera setting.

Iskakov~\etal~\cite{iskakov2019learnable} proposed to learn 3D pose directly via differentiable triangulation~\cite{hartley2003multiple}.
One key difference between their learnable triangulation and ours is that Iskakov~\etal~\cite{iskakov2019learnable} fuses features with 3D voxel feature maps,
which is more computationally expensive and memory intensive than our method that fuses 3D-aware features in 2D feature maps (\tableref{tab:human_extra}). 

\paragraph{Multi-view Hand Pose Estimation:} Most 3D hand pose estimation works focus on either monocular RGB images or monocular depth images~\cite{yuan2017bighand2,yuan2018rgb,xiang2019monocular,dibra2018monocular,zhang2019end,iqbal2018hand,wu2018handmap,zhou2018hbe,ge2018point,cai2018weakly,li2019point,yang2019disentangling,ye2018occlusion}. 
%Boukhayma~\etal~\cite{boukhayma20193d} created a synthetic dataset of paired hand images with their groundtruth camera and hand parameters. 
In contrast, there are fewer works on multi-view 3D hand pose estimation, especially two-hand pose estimation due to the difficulty of obtaining  multi-view hand data annotations. Simon~\etal~\cite{simon2017hand} proposed to iteratively boost single image 2D hand keypoint detection performance using multi-view bootstrapping. Garcia~\etal~\cite{garcia2018first} introduced a first-person two-hand action dataset with RGB-D and 3D hand pose annotations. Unfortunately, only the right hand is annotated. To showcase our \ours\ on the application of multi-view two-hand pose estimation, we use our internal \handdataset \ dataset. 

\paragraph{Epipolar Geometry in Deep Neural Networks:} Prasad~\etal~\cite{prasad2018epipolar} applied epipolar constraints to depth regression with the essential matrix. Yang~\etal~\cite{yang2019learning} proposed to use symmetric epipolar distance for data-adaptive interest points. MONET~\cite{jafarian2018monet} used epipolar divergence for multi-view semi-supervised keypoint detection.  Different from the above methods, we leverage the epipolar geometry for deep feature fusion.

\paragraph{Attention Mechanism:} Vaswani~\etal~\cite{vaswani2017attention} first proposed a transformer for sequence modeling based solely on attention mechanisms. Non-local networks ~\cite{wang2018non} were introduced for capturing long-term dependencies in videos for video classification. Our approach is named epipolar attention, because we compute attention weights along the epipolar line based on feature similarity and use these weights to fuse features.

%%%%%%%%%%%%%%%%%%%%%%%%%%%%%%%%%%%%%%%%%%%%%%%%%%%%%%%% Approach %%%%%%%%%%%%%%%%%%%%%%%%%%%%%%%%%%%%%%%%%%%%%%%%%%%%%%%
\section{The \OURS}\label{sec:approach}

Our \ours \  consists of two main components: the epipolar sampler and the feature fusion module.
Given a point $p$ in the reference view,
the epipolar sampler will 
sample features along the corresponding epipolar line in the source view.
The feature fusion module will then take (1) all the features at the sampled locations in the source view
and (2) the feature at $p$ in the reference view to produce a final 3D-aware feature.
Note that this is done densely for all locations of an intermediate feature map from the reference view.
We now provide the details to the two components.
%, as well as some implementation details on how to handle image transformations while using the \ours.

\subsection{The Epipolar Sampler}
%\paragraph{Notations:}
We first define the notations used to describe the epipolar sampler.
Given two images captured at the same time but from different views, namely, \mainview\ $\mainimage$\ and \otherview\ $\otherimage$,  we denote their projection matrices as $\mainproj,\,\otherproj \in \mathbb{R}^{3\times4}$ and camera centers as $\maincameracenter, \, \othercameracenter \in \mathbb{R}^4$ in homogeneous coordinates.
As illustrated in \figref{fig:arch}, assuming the camera centers  do not overlap, the epipolar line $\otherepipolarline$ corresponding to a given query pixel $\mainloc = (x, y, 1)$ in $\mainimage$ can be deterministically located on $\otherimage$ as follows \cite{hartley2003multiple}. 
% on the \mainview\ is a function of the corresponding position of point $P$ in 3D space.
\begin{align}
\otherepipolarline = [\otherproj\maincameracenter]_\times \otherproj \mainproj^+ \mainloc,
\end{align}
where $\mainproj^+$ is the pseudo-inverse of $\mainproj$, and $[\cdot]_\times$ represents the skew symmetric matrix. $\mainloc$'s corresponding point in the \otherview : $\otherloc$, should lie on the epipolar line: $\otherepipolarline^T\otherloc=0$.
%The projection of $P$ onto the \otherview\ $\otherimage$, denoted as 

%\subsection{Formulation}

%\paragraph{Epipolar sampler:} 
%We define the \epipolarsampler\ on a pair of views. %, that searches the correct position along the epipolar line. % fuse features from a \otherview.
Given the epipolar line $\otherepipolarline$ of the \otherview,
the \epipolarsampler\ uniformly samples $\samplesize$ locations (64 in our experiments) along the visible portion of the epipolar line, i.e., the intersection of $\otherimage$ and $\otherepipolarline$. 
The sampled locations form a set $\mathcal{P}'$ with cardinality $K$.
%$\sampler$ uniformly samples $\samplesize$ locations (\eg, 64) on the , thus forming a set $\mathcal{P}'$ of cardinality $K$.
The \epipolarsampler\ samples sub-pixel locations (real value coordinates) via bilinear interpolation. 
For query points whose epipolar lines do not intersect with $\otherimage$ at all, we simply skip them.
Please also see supplementary materials for details on how to handle image transformations for the \ours.

%The function takes as input the query location $\mainloc$ on the \mainview, and the projection matrices $\mainproj, \otherproj$ as shown below. 
%\begin{align}\label{eq:sampler}
%\mathcal{P}' = \sampler_K(\mainloc, \mainproj, \otherproj)
%\end{align}

%The output of the \epipolarsampler$\,$ are the location of the sampled points along the epipolar line in the \otherview\, .

\subsection{Feature Fusion Module}
Ideally, if we knew the ground-truth $\otherloc$ in the source view that corresponds to $p$ in the reference view, 
then all we need to do is sample the feature at $\otherloc$ in the source view: $F_{\othernotationabbr}(\otherloc)$, 
and then combine it with the feature at $\mainloc$ in the reference view: $F_{\mainnotationabbr}(\mainloc)$.
However, we do not know the ground-truth $\otherloc$.
Therefore, inspired by Transformer~\cite{vaswani2017attention} and non-local networks ~\cite{wang2018non},
we approximate $F_{\othernotationabbr}(\otherloc)$ 
by a weighted sum of all the features along the epipolar line as follows:
\begin{align}
\overline{F}_{\othernotationabbr}(p) = \sum_{\otherloc \in \mathcal{P}'}\text{sim}(\mainloc, \otherloc) F_{\othernotationabbr}(\otherloc)
% \text{softmax}( x_{\mainloc}^T \sampledfeat_{\mainloc}) \sampledfeat_{\mainloc}
\label{eq:otherfeat}
\end{align}
where the pairwise function $\text{sim}(\cdot, \cdot)$ computes the similarity score between two vectors.
More specifically, it is the dot-product followed by a softmax function.
%where $F_{\othernotation}(\cdot)$ is the feature vector at a given location.
%The weights are computed by the similarity between features, i.e., 
%, we compute the similarity between the feature at $\mainloc$ and points in $\mathcal{P}'$ using dot-product and softmax. Using the similarity scores as the weights, the sampled features from \otherview\ $F_{\othernotation}$ is thus summarized by the weighted sum of the features on each point :

\newcommand{\archidentity}{Identity Gaussian}
\newcommand{\archembedded}{Bottleneck Embedded Gaussian}

\begin{figure}
\centering
    % \begin{center}
    % \includegraphics[width=1.1\linewidth]{imgs/module.pdf}
    \includegraphics[width=\linewidth]{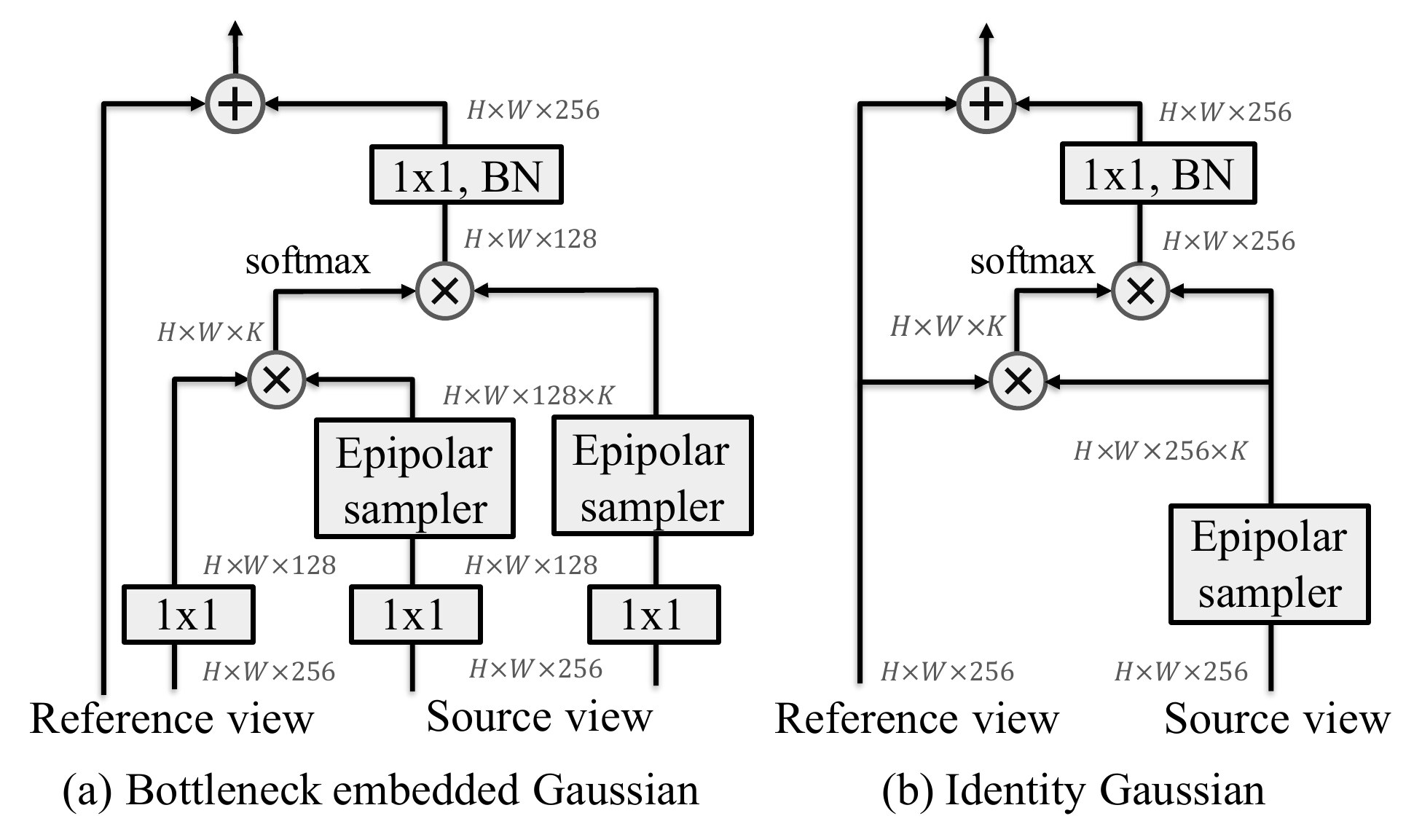}
    \caption{Different feature fusion module architectures. The feature maps are shown along with the shape of their tensors, \eg, $H\times W\times 256$ for 256 channels.  "$\oplus$" denotes the element-wise sum and "$\otimes$" denotes the batch matrix multiplication where the batch size is  $H\times W$. }
    \label{fig:module}
    % \end{center}
\end{figure}

Once we have the feature from the \otherview: $\overline{F}_{\othernotationabbr}(\mainloc)$,
we now need to fuse it with the feature in the \mainview: $F_{\mainnotationabbr}(p)$.
One straightforward way to fuse the features is motivated by the residual block \cite{resnet}, where the feature from the \otherview\ goes through a transformation  $\weightz$ before being added to the features of the \mainview\ as shown in \figref{fig:module}~(b) and the following equation:
%by adding the transformed $F_{\othernotation}$ with the original $F_{\mainnotation}$ as follows:
\begin{align}\label{eq:fuse}
    F_{\textrm{fused}}(\mainloc) = F_{\mainnotationabbr}(\mainloc) + \weightz(\overline{F}_{\othernotationabbr}(\mainloc))
\end{align}
The weights $\weightz$ are $1\times 1$ convolutions in our experiments. 
%In this case, keeping a copy of the original $F_{\mainnotationabbr}$ feature is similar to the design of the residual block. 
We refer to this method as the \textit{\archidentity} architecture.
Note that the output $F_{\textrm{fused}}$ is of the same shape as the input $F_{\mainnotationabbr}$, thus this property enables us to insert the \ours$\,$ module into different stages of many existing networks.  

We also explore the \textit{\archembedded} architecture, which was popularized by non-local networks~\cite{wang2018non}, as the feature fusion module as shown in \figref{fig:module}~(a). 
Before the \ours, the features from the \mainview\ and \otherview\ goes through an embedded Gaussian kernel, where the channel size is down-sampled by a factor of two, and the output is up-sampled back, so that the shape of the fused feature still matches the input's shape. 

%%%%%%%%%%%%%%%%%%%%%%%%%%%%%%%%%%%%%%%%%%%%%%%%%%%%%%%% Experiments %%%%%%%%%%%%%%%%%%%%%%%%%%%%%%%%%%%%%%%%%%%%%%%%%%%%%%%

\section{Experiments}
We have conducted our experiments on two large-scale pose estimation datasets with multi-view images and ground-truth 3D pose annotations: an internal hand dataset \handdataset, and a publicly available human pose dataset \humandataset. 
%For fair comparison with the baseline detectors, unless specified, additional data augmentation is not used.
%technique mentioned in Section~\ref{sec:augmentation} is not used.

\paragraph{\handdataset\ dataset:}

\handdataset\ is an internal hand dataset that is captured in a synchronized multi-view studio with 34 color cameras and 46 monochrome cameras. Only the color cameras were used in our experiments.
We captured 23 subjects doing various one and two handed poses.
We then annotated the 3D location of the hand for 7.6K unique timestamps, which led to 257K annotated 2D hands when we projected the 3D annotations to all 34 2D views. 248K images were used for training, and 9K images were used for testing.
%The camera distribution is shown in \figref{fig:cam_distribution}.
For each hand, 21 keypoints were annotated, so there are 42 unique points for two hands.
%Each recording consists of 40 left, 40 right and 15 interacting hand sequences, including two types of motion: free moves with minimal instructions, and short transitions from neutral pose to a pre-defined hand pose then transition back to the neutral pose. For each hand, 21 keypoints are annotated and therefore there are 42 unique points for the two hands.

% \begin{figure}
%     % \begin{center}
%     \includegraphics[width=0.8\linewidth]{imgs/distribution.png}
%     \caption{Studio used to capture \handdataset.}
%     \label{fig:cam_distribution}
%     % \end{center}
% \end{figure}

\paragraph{\humandataset :} 

\humandataset\  is one of the largest 3D human pose benchmarks captured with four cameras and has 3.6M 3D annotations available. The cameras are located at the corners of a rectangular room and therefore have larger baselines. This leads to a major difference compared to \handdataset\ -- the viewing angle difference between the cameras in \humandataset\ are significantly larger.

\paragraph{Evaluation metric: } 
% For the 2D pose estimation, we used mean square error distance \todo{final check} and the area under the curve (AUC) of the percentage of correct keypoints to evaluate the accuracy of the 2D keypoint detection.
During training, we use Mean Squared Error (MSE) between the predicted and ground truth heatmaps as the loss.  A ground truth heatmap is generated by applying a 2D Gaussian centered on a joint. To estimate the accuracy of 3D pose prediction, we adopt the MPJPE (Mean Per Joint Position Error) metric. It is one of the most popular evaluation metrics, which is referred to as Protocol $\#1$ in ~\cite{martinez2017simple}. It is calculated by the average of the L2 distance between the ground-truth and predictions of each joint.

\subsection{Ablation study on \handdataset\ Dataset}
We have performed a series of ablation studies on the \handdataset\ dataset to better understand the \ours, and at the same time to understand the effect of
different design choices.

\begin{figure}
    \centering
    \includegraphics[width=0.9\linewidth]{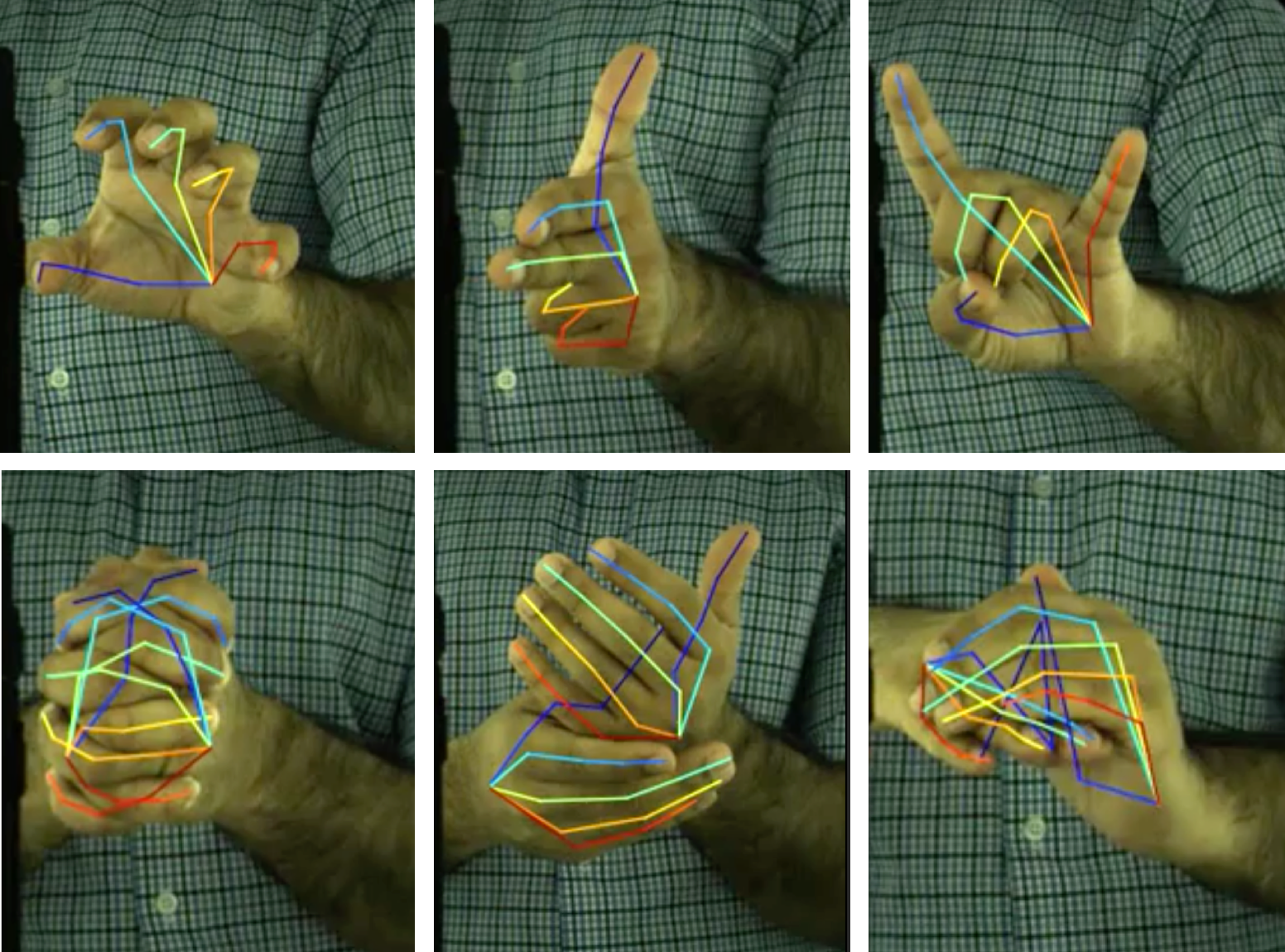}
    \caption{
    Visualization of predictions for \handdataset. The images are cropped for ease of visualization. Our model still fails on challenging hand poses with occlusions (bottom right).
    %For \handdataset\ dataset, here is a visualization of the groundtruth and the predictions by baseline detector and our \ours\ detector. Note the pinky finger of the right hand is occluded by a black stand but the model can still estimate a position with information from other views. 
    }
    \label{fig:hand_result}
    % \end{center}
\end{figure}

%\todo{Check the consistency of validation and training set}
We trained a single-stage Hourglass network~\cite{newell2016stacked} with the \ours\ included. %\ plugged in.% following~\cite{newell2016stacked}.
To predict the 3D joint positions, we run the 2D detector trained with the \ours\ on all the views, and then perform triangulation with RANSAC to get the final 3D joint locations.
During prediction time, our 2D detector requires features from the \otherview, which is randomly selected from the pool of cameras that were used as the \otherview\ during training. 
We downsampled the images to %5 frames per second with 
resolution 512$\times$336 for training and testing. 
%, and only use the images from the 34 RGB cameras for our experiments.
\figref{fig:hand_result} visualizes some predictions from our model.
%we show a visualization of the joints predicted by the baseline detector and the detector with our epipolar transformer.

\paragraph{Feature fusion module design:}
First, we compare the \archembedded\ ((\figref{fig:module}~(a)) popularized in the non-local networks~\cite{wang2018non} with \archidentity\ (\figref{fig:module}~(b)). 
As shown in \tableref{tab:module}, \archidentity\ performs slightly better. 
We hypothesize that, unlike video classification~\cite{wang2018non}, pose estimation needs accurate correspondences, 
so down-sampling in \archembedded\ could be detrimental to the performance.
%which lead to \archidentity\ outperforming \archembedded. 
%Also, the correspondences might be more accurate without the convolution transformation in \archembedded. 

\paragraph{Max or softmax?}
We use softmax to obtain the weights along the epipolar line. 
Another option is to use max because our goal is to find a single ``correct'' point on the epipolar line. 
Shown in \tableref{tab:module}, max performs slightly worse than softmax. 
We hypothesize that it is because softmax produces gradients for all samples on the epipolar line, which helps training. %, i.e., when the correspondence is incorrect, wrong features would be fused to the \mainview\ with some penalty. 

\begin{table*}
  \adjustbox{valign=t}{
  \begin{minipage}[!t]{0.42\linewidth}
    %   \centering
    % \begin{center}
    \setlength\tabcolsep{2.5pt}
    \begin{tabular}{l|cc}
        \hline
        \multirow{2}{*}{Architecture design} & \multicolumn{2}{c}{MPJPE (mm)} \\ \cline{2-3} 
         & \handdataset & H3.6M \\ \hline
        \archembedded & 4.99 & 35.7 \\ \hline
        \archidentity\ + max &  4.97 & - \\ \hline
        \archidentity\ + softmax & \textbf{\handour36view} & \textbf{\humansmallresnetfiftydlt} \\ \hline
    \end{tabular}
    \caption{Architecture design comparison for both \handdataset\ and \humandataset. }
    \label{tab:module}
  \end{minipage}
}
  \adjustbox{valign=t}{
      \begin{minipage}[!t]{0.25\linewidth}
      \centering
    % \begin{center}
    \begin{tabular}{l|c}
    \hline
    Stage & MPJPE (mm)      \\ \hline
    early + late     &  5.03 \\ %\hline
    early         &  4.96  \\ %\hline
    late     & \textbf{\handour36view} \\ \hline
    \end{tabular}
    \caption{\Ours\  plugged into different stages of Hourglass networks~\cite{newell2016stacked} on \handdataset\ dataset.}
    \label{tab:hand_stage}
    % \end{center}
  \end{minipage}
}
\adjustbox{valign=t}{
 \begin{minipage}[!t]{0.25\linewidth}
  \centering
    % \begin{center}
    \begin{tabular}{l|c}
    \hline
    Inference & MPJPE (mm)            \\ \hline
    baseline     & \handbaseline36view          \\ \hline
    single source view    & \handour36view \\ 
    multi source views & \textbf{4.83} \\ \hline
    \end{tabular}
    \caption{Different number of neighboring source views for inference on \handdataset.}
    \label{tab:hand_inference}
% \end{center}
  \end{minipage}}%\hfill
\end{table*}

% \begin{table}[]
% \begin{center}
% \begin{tabular}{l|c}
% \hline
% source camera & MPJPE            \\ \hline
% dot-product + max     &     4.97       \\ \hline
% embeded Gaussian      & 4.99          \\ \hline
% Gaussian     & \textbf{\handour36view} \\ \hline
% \end{tabular}
% \caption{Different \ours\ added to Hourglass network on \handdataset}
% \label{tab:module}
% \end{center}
% \end{table}

\newcommand{\rotationsetting}{6$^\circ$, 12$^\circ$, 24$^\circ$ and 42$^\circ$}
\newcommand{\rotationbest}{24$^\circ$}
\newcommand{\rotation}{viewing angle}
\newcommand{\Rotation}{viewing angle}
\paragraph{Viewing angle:}

\begin{figure*}
  \begin{minipage}[c]{0.7\linewidth}
  \centering
    % \begin{center}
    \includegraphics[width=\linewidth]{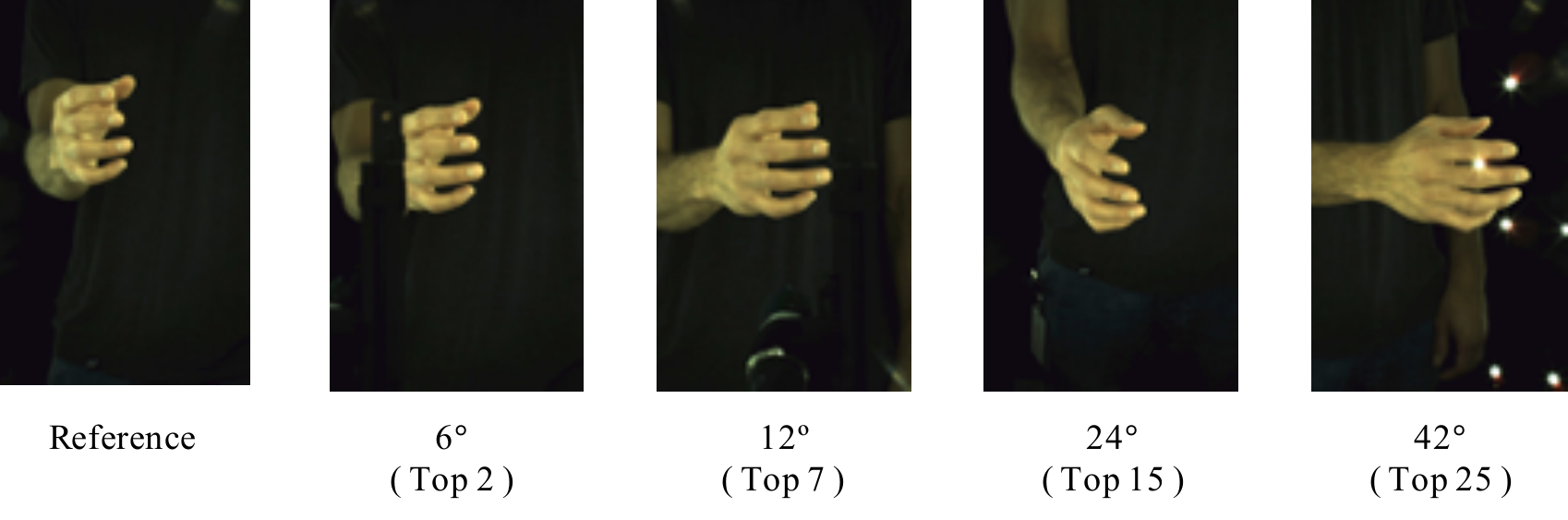}
    % \end{center}
  \end{minipage}
  \hspace{0.1cm}
  \begin{minipage}[c]{0.3\linewidth}
%   \centering
    % \begin{center}
    \includegraphics[width=0.9\linewidth]{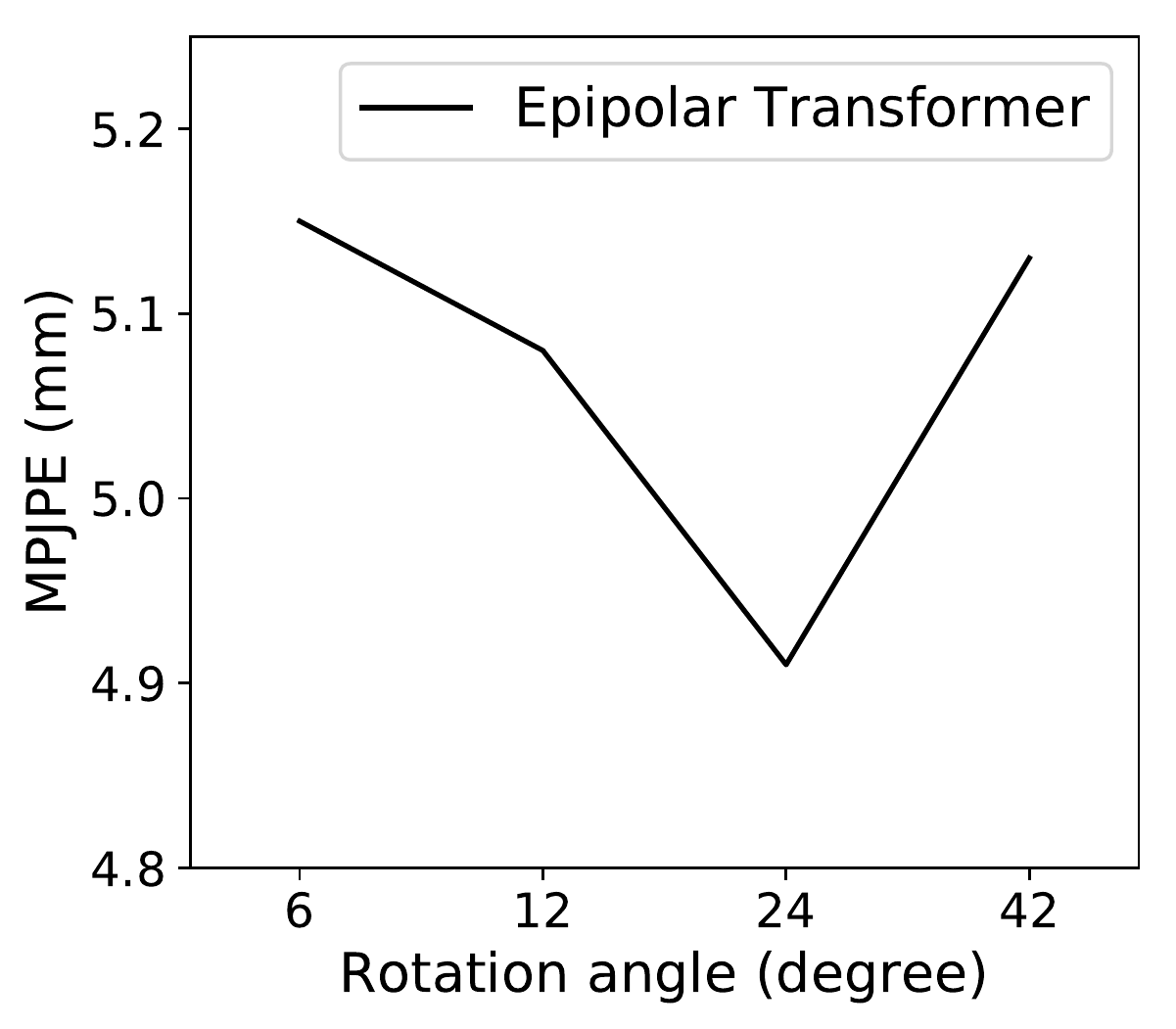}
% \begin{tabular}{l|c}
% \hline
% source camera & MPJPE            \\ \hline
% baseline     & \handbaseline36view          \\ \hline
% top 5         & 5.15          \\
% top 5-10      & 5.08          \\ 
% top 10-20     & \textbf{\handour36view} \\
% top 20-30     & 5.13          \\ \hline
% \end{tabular}
    % \end{center}
    \end{minipage}
\caption{Illustration of different \rotation s between the reference and source view. From the left to the right are the reference view image, and 
the images with \rotation s difference \rotationsetting\ respectively.
%its 2$^{th}$ / 7$^{th}$ / 15$^{th}$ / 25$^{th}$ nearest neighbours accordingly.
On the right is the performance measured in MPJPE under different \rotation s for \handdataset.}
\label{fig:baseline}
\end{figure*}
% \begin{table}[]
% \centering
% \begin{tabular}{l|c}
% \hline
% source camera & MPJPE            \\ \hline
% no fusion     & \handbaseline36view          \\ \hline
% top 5         & 5.15          \\ \hline
% top 5-10      & 5.08          \\ \hline
% top 10-20     & \textbf{\handour36view} \\ \hline
% top 20-30     & 5.13          \\ \hline
% \end{tabular}
% \caption{}
% \label{tab:my-table}
% \end{table}

We now study how the \rotation\ difference of the selected neighboring camera affects the performance of the \ours. The intuition is that, if the \otherview\  has a viewing angle very similar to the \mainview, the features from the two views may be too similar, thus not very informative for our fusion module. 
However, if the views are too far apart, the feature matching becomes harder. There is also a higher chance that a point is occluded in one of the views, which makes matching even harder. Therefore, we experiment with four viewing angle settings during training: \rotationsetting. 
The \otherview\ is randomly selected from ten cameras whose viewing angles are closest to the chosen \rotation.
\figref{fig:baseline} shows examples of the \otherview s with  different viewing angles from an example \mainview. As also shown in \figref{fig:baseline}, the \ours\ is most effective around viewing angle difference \rotationbest, which is the setting we use by default for other \handdataset\   experiments.

%When the \mainview\ and \otherview\ are too close, the features are so similar that feature fusion does not help that much. When the \mainview\ and the \otherview\ are too far away from each other, the features are quite different, which makes finding correspondences difficult. 

% \paragraph{RGB feature and deep feature matching.}

% In order to prove the effectiveness of matching deep features, given the same epipolar constraint, we compare two matching criteria: matching based on RGB feature and matching based on deep feature. In the epipolar transformer, we searched along the epipolar line to find the location with the most similar deep feature, and merge the deep feature accordingly. In the RGB feature matching setting, we merged the deep feature according to the correspondence based on the RGB feature (3 channel feature) matching. As shown in \ref{tab:matching_criterion}, matching based on RGB feature performs worse than matching with the more robust and high-level deep feature. 

% \begin{table}[]
% \begin{center}
% \begin{tabular}{l|c}
% \hline
% Matching criterion & MPJPE            \\ \hline
% RGB feature    & 5.12668          \\ \hline
% deep feature         & \textbf{\handour36view}          \\ \hline
% \end{tabular}
% \caption{Matching based on RGB feature or deep feature}
% \label{tab:matching_criterion}
% \end{center}
% \end{table}

\paragraph{Which stage to insert the \ours:} 
We perform experiments to test the ideal location for inserting the \ours\ into a network.
%An interesting question is: when should add the \ours\ and fuse the features?
We tested two settings: ``late'' means the \ours\ is inserted before the final prediction layer, and ``early'' means we insert the module before the Hourglass unit~\cite{newell2016stacked}. The exact locations are detailed in the supplementary materials.
As shown in \tableref{tab:hand_stage}, there is no significant difference where we add the \ours. 
%it is best to add it in the late stage. This observation is different from non-local network~\cite{wang2018non} which uses ResNet~\cite{resnet}. One possible explanation is that the later stages of the hourglass network retain higher resolution than ResNet. Thus more information is retained, and doing matching is better then. Also, fusing twice with \ours\ does not improve the performance. 
In the rest of this paper, we fuse at the late stage by default.

% \begin{table}[]
% \begin{center}
% \begin{tabular}{l|c}
% \hline
% stage & MPJPE            \\ \hline
% early + late     & 5.03 \\ \hline
% early         & 4.96          \\ \hline
% late     & \textbf{\handour36view} \\ \hline
% \end{tabular}
% \caption{\Ours\ added into different stages of hourglass network~\cite{newell2016stacked}}
% \label{tab:hand_stage}
% \end{center}
% \end{table}

\begin{figure}
\centering
% \begin{center}
    % \includegraphics[width=0.8\linewidth]{imgs/nviews_top10-20.pdf}
    % \includegraphics[width=1.0\linewidth]{imgs/nviews_top10-20.pdf}
    \includegraphics[width=0.9\linewidth]{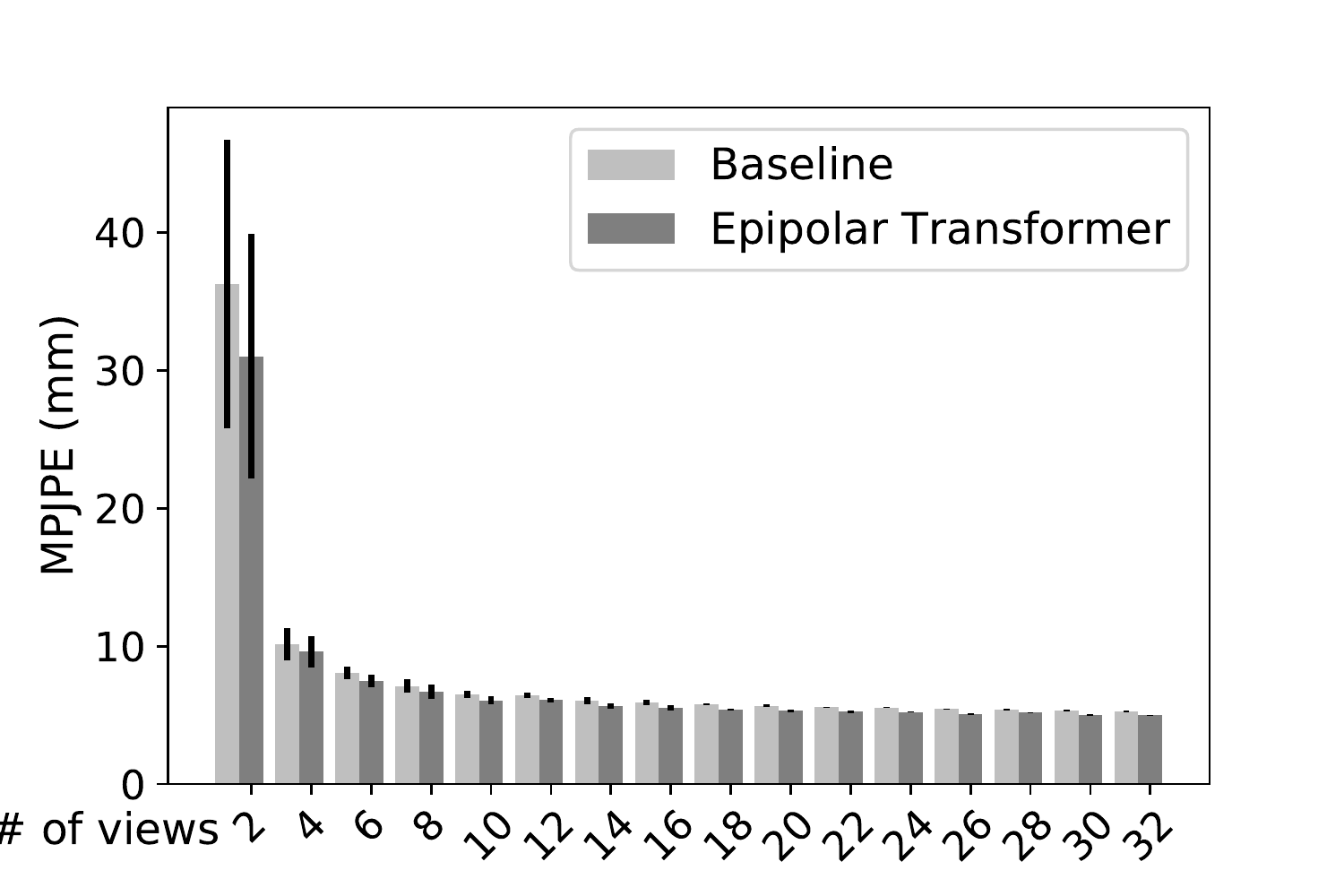}
    
    \caption{MPJPE by varying the number of views used for prediction on \handdataset . The black lines indicate the standard deviation.}
    \label{fig:nviews_handsy}
% \end{center}
\end{figure}

\paragraph{Effect of the number of views used during test time:} 
%In the last paragraph we find that the feature is well-learned if the \rotation\ is about \rotationbest. \Ours\ can learn better feature with this \rotation. However, 
We explore how the number of views used during test time affects the final performance.
Since there are many different combinations to sample reference and source views, 
we randomly sample views multiple times (up to 100 times when there are few cameras) and ensure that for each camera there is at least one neighboring camera with viewing angle difference near \rotationbest.  
%To study the effectiveness of our method with different number of the cameras, 
The baseline compared is a vanilla Hourglass network without using the \ours.
As shown in \figref{fig:nviews_handsy}, the network trained with the \ours\ consistently outperforms the baseline. 
When very few views are used (\eg, two views), the relative improvement using the \ours\ is around 15\%. This supports our argument: \ours s enable the network to obtain better 2D keypoints using information from neighboring views, and the information is crucial when there are fewer views. Even with more views, the \ours\ is still able to improve upon the baseline by around 10\%.
%If there are enough cameras, it is enough to triangulate a decent 3D locations with RANSAC. 

% \begin{table}[]
% \begin{center}
% \begin{tabular}{l|c|c}
% \hline
%          & baseline & ours              \\ \hline
% 2 views  & 45.09 & \textbf{30.47} \\ \hline
% 20 views & 5.67  & \textbf{5.17}  \\ \hline
% 34 views & \handbaseline36view  & \textbf{\handour36view}  \\ \hline
% \end{tabular}
% \end{center}
% \caption{Number of views on \handdataset}
% \label{tab:nviews}
% \end{table}

\paragraph{Inference with multiple source views:} The \ours\ we introduced is not limited to fusing features from two views. During testing, we can choose different neighboring views as the \otherview\ and select the prediction with the highest confidence (\ie, highest peak on the heatmap). As shown in \tableref{tab:hand_inference}, we run prediction with ten different neighboring views and select the prediction with the highest confidence for each individual joint. Testing with multi-views reduces the MPJPE error only by 0.1 mm, which is insignificant. The performance might be further improved when training with more than two views, like in MVSNet~\cite{yao2018mvsnet}.

% \begin{table}[]
% \begin{center}
% \begin{tabular}{l|c}
% \hline
% inference & MPJPE            \\ \hline
% baseline     & \handbaseline36view          \\ \hline
% one source view    & \handour36view \\ 
% multi source view & \textbf{4.83} \\ \hline
% \end{tabular}
% \caption{Multiview inference using \ours}
% \label{tab:hand_inference}
% \end{center}
% \end{table}

\paragraph{Comparison with \crossviewfusion:} Qiu\etalcite{qiu2019cross} learns fixed global attention weights for each pair of views. A limitation of this method is it requires more images per view to learn the attention weights. 
On \handdataset, \crossviewfusion\ performs even worse than the baseline detector as shown in \tableref{tab:hand_fusion}. One likely reason is because there are only about 3K images per view on \handdataset, instead of 312K images per view on \humandataset. Besides, \humandataset\ only has four views, which makes it easier to learn the pairwise attention weights, but learning the weights for \handdatasetrgbviews\ views for \handdataset\  is significantly harder. 

\begin{table}[t]
\begin{tabular}{l|cc}
\hline
\multirow{2}{*}{} & \multicolumn{2}{c}{MPJPE (mm)} \\ \cline{2-3} 
 & \begin{tabular}[c]{@{}c@{}}\handdataset\\ 7k imgs/view\end{tabular} & \begin{tabular}[c]{@{}c@{}}Human3.6M\\ 312k imgs/view\end{tabular} \\ \hline
\crossviewfusion & 6.29 & 45.47 \\ \hline
baseline & 5.46 & 48.73 \\ \hline
\ours & \textbf{\handour36view} & \textbf{\humansmallresnetfiftydlt} \\ \hline
\end{tabular}
\caption{Comparison with \crossviewfusion. 
The baseline is using Hourglass networks~\cite{newell2016stacked} for \handdataset\  and Resnet-50~\cite{resnet} for \humandataset\  without view fusion.}
\label{tab:hand_fusion}
\end{table}

% \begin{table}[]
% \begin{center}
% \begin{tabular}{l|c}
% \hline
% fusion method & MPJPE            \\ \hline
% \crossviewfusion     & 6.29          \\ \hline
% no fusion         & 5.15          \\ \hline
% \ours     & \textbf{\handour36view} \\ \hline
% \end{tabular}
% \caption{different fusion methods comparison}
% \label{tab:hand_fusion}
% \end{center}
% \end{table}

\subsection{Human3.6M Dataset}
We conducted experiments on the publicly available \humandataset\ dataset. We adopt the same training and testing sets as in \cite{qiu2019cross}, where subjects 1, 5, 6, 7, 8 are used for training, and 9, 11 are for testing. 

As there are only four views in \humandataset, we choose a closest view as the \otherview. We adopt ResNet-50 with image resolution 256$\times$256 proposed in simple baselines for human pose estimation~\cite{xiao2018simple} as our backbone network. We use the ImageNet~\cite{imagenet} pre-trained model~\cite{pytorch} for initialization. The networks are trained for 20 epochs with batch size 16 and Adam optimizer~\cite{kingma2014adam}. Learning rate decays were set at 10 and 15 epochs. Unless specified, we follow Qiu\etalcite{qiu2019cross}'s setting and do not do additional data augmentation for a fair comparison. We also follow Qiu\etalcite{qiu2019cross} for other hyper-parameters. Following \cite{qiu2019cross}, as there are only four cameras in this dataset, direct linear transformation (DLT) is used for triangulation (Hartley \& Zisserman~\cite{hartley2003multiple}, p.312), instead of RANSAC. % which also needs tuning the inlier/outlier threshold. 
%Well-tuned RANSAC is slightly better than DLT~\cite{iskakov2019learnable}. 

\newcommand{\crossviewsum}{sum over epipolar line}
\newcommand{\crossviewmax}{max over epipolar line}
\paragraph{2D Pose Estimation:}  Again following Qiu\etalcite{qiu2019cross}, the 2D pose estimation accuracy is measured by Joint Detection Rate (JDR), which measures the percentage of the successfully detected joints. A joint is detected if the distance between the estimated location and the ground truth is smaller than half of the head size~\cite{andriluka20142d}. 

2D pose estimation results are shown in \tableref{tab:human_jdr}. As shown in Qiu\etalcite{qiu2019cross}, one way to compute a cross-view score for a specific reference view location is to do a sum or max of the heatmap prediction scores along its corresponding epipolar line in the source view, but this does not lead to good performance. So \crossviewfusion\ improved performance by fusing with learned global attention. %, which resembles the non-local networks~\cite{wang2018non}. 
In contrast, the \ours\ neither operates on heatmap prediction scores nor does a global fusion. It attends to the \textit{intermediate features locally along the epipolar line}. Using the same backbone ResNet-50 with input image size 256$\times$256, the model with \ours\ achieves \textbf{\humansmallresnetfiftydltjdr\%} JDR, which outperforms 95.9\% JDR from Qiu\etalcite{qiu2019cross} by 1\%. The improvement suggests that fusing along the epipolar line is better than fusing globally. 
We further apply data augmentation which consists of random scales drawn from a truncated normal distribution
%\footnote{https://github.com/microsoft/human-pose-estimation.pytorch}
$\text{TN}(1, 0.25^2, 0.75, 1.25)$ and random rotations from $\text{TN}(0^{\circ}, (30^{\circ})^2, -60^{\circ}, 60^{\circ})$~\cite{xiao2018simple}. JDR is further improved to \textbf{\humansmallresnetfiftydltaugjdr\%} JDR.

\begin{table*}[]
\setlength\tabcolsep{2.5pt}
\begin{tabular}{l|l|l|llllllllllll}
    \toprule
 & Net & scale & shlder & elb & wri & hip & knee & ankle & root & belly & neck & nose & head & Avg \\ \hline
- & R152 & 320 & 88.50 & 88.94 & 85.72 & 90.37 & 94.04 & 90.11 & - & - & - & - & - & - \\
\crossviewsum\cite{qiu2019cross} & R152 & 320 & 91.36 & 91.23 & 89.63 & 96.19 & 94.14 & 90.38 & - & - & - & - & - & - \\
\crossviewmax\cite{qiu2019cross} & R152 & 320 & 92.67 & 92.45 & 91.57 & 97.69 & 95.01 & 91.88 & - & - & - & - & - & - \\
\crossviewfusion & R152 & 320 & 95.58 & 95.83 & \textbf{95.01} & 99.36 & 97.96 & 94.75 & - & - & - & - & - & - \\
\crossviewfusion$^\star$ & R50 & 320 & 95.6 & 95.0 & 93.7 & 96.6 & 95.5 & 92.8 & 96.7 & 96.4 & 96.5 & 96.4 & 96.2 & 95.9 \\
\crossviewfusion$^\star$ & R50 & 256 & 86.1 & 86.5 & 82.4 & 96.7 & 91.5 & 79.0 & \textbf{100.0} & 94.1 & 93.7 & 95.4 & 95.5 & 95.1 \\ \hline
\ours & R50 & 256 & 96.44 & 94.16 & 92.16 & 98.95 & 97.26 & 96.62 & 99.89 & 99.86 & 99.68 & \textbf{99.78} & \textbf{99.63} & \humansmallresnetfiftydltjdr \\
\textbf{\ours$^+$} & R50 & 256 & \textbf{97.71} & \textbf{97.34} & 94.85 & \textbf{99.77} & \textbf{98.32} & \textbf{97.55} & 99.99 & \textbf{99.99} & \textbf{99.76} & 99.74 & 99.54 & \textbf{\humansmallresnetfiftydltaugjdr} \\ \bottomrule
\end{tabular}
\caption[]{2D pose estimation accuracy comparison on \humandataset\  where no external training data is used unless specified. The metric is joint detection rate, JDR (\%). $+$: indicates using data augmentation.
%mentioned in Section~\ref{sec:augmentation}.
"-": We cite numbers from \cite{qiu2019cross} and these entries were absent. $\star$: We trained the models using released code~\cite{qiu2019cross}. %\footnotemark. 
R50 and R152 are ResNet-50 and ResNet-152~\cite{resnet} respectively. Scale is the input resolution of the network.}
\label{tab:human_jdr}
\end{table*}
% \footnotetext{\label{note:fuse}github.com/microsoft/multiview-human-pose-estimation-pytorch}

\paragraph{Visualization of feature-matching:}

\begin{figure}
    \includegraphics[width=\linewidth]{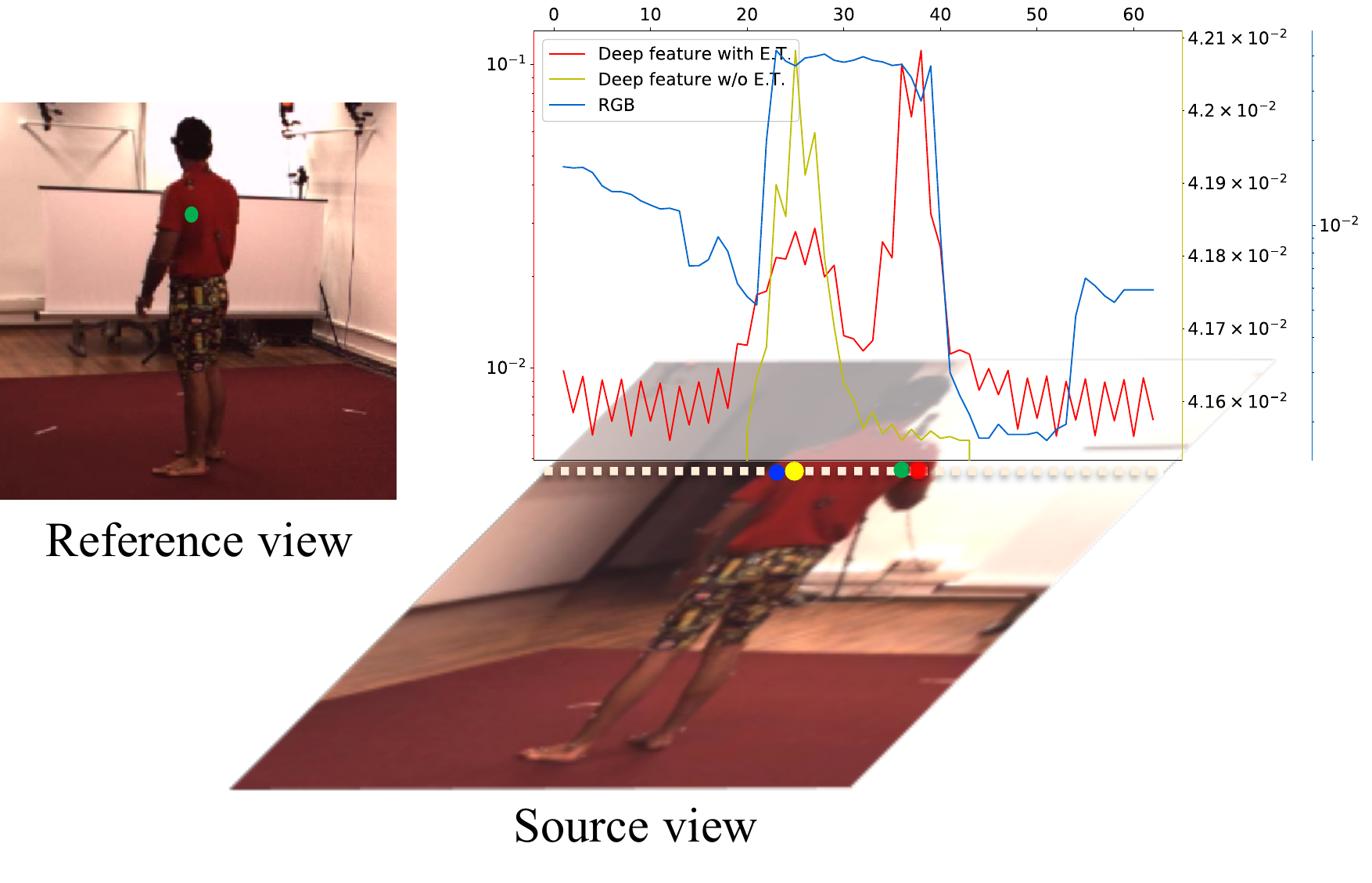}
\caption{Visualization of the matching results along the epipolar line for various features on \humandataset. The (occluded) right wrist is selected and denoted by a green dot in the reference view. Features used for matching are (a) deep features learned through the Epipolar Transformer (deep features with E.T.), (b) deep feature learned by ResNet-50~\cite{resnet} without epipolar transformer (deep features w/o E.T.), and (c) color features (specifically RGB converted to LAB and then excluding the L channel). Green dot on the source view is the corresponding point of the ground-truth. %(\textit{best viewed in color})
}

\label{fig:matching-comp-hm36-main-body}
\end{figure}

The main advantage of the \ours\ is that it is easily interpretable through visualizing the feature-matching similarity score along the epipolar line.
We visualize the results of feature-matching along the epipolar line for color features, deep features learned without the \ours, and the features learned through the \ours.
For the color features, we first convert the RGB image to the LAB color space. Then we discard the L-channel and only use the AB channel to be more invariant to light intensity.
\figref{fig:matching-comp-hm36-main-body} shows a challenging example where the joint-of-interest is totally occluded in both views. However, given that the features learned with the \ours\  have access to multi-view information in the 2D detector itself, the matching along the epipolar line finds the ``semantically correct'' position, i.e., still finds the occluded right wrist, which is the desired behavior for a pose detector. However, the features without the awareness of the multi-view information have the highest matching score at the ``physically correct'' location, which is still correct in terms of finding correspondences, but not as useful to reason about occlusions for occluded joints. More examples are shown in the supplementary material.

\begin{figure}
\centering
% \begin{center}
    \includegraphics[width=0.9\linewidth]{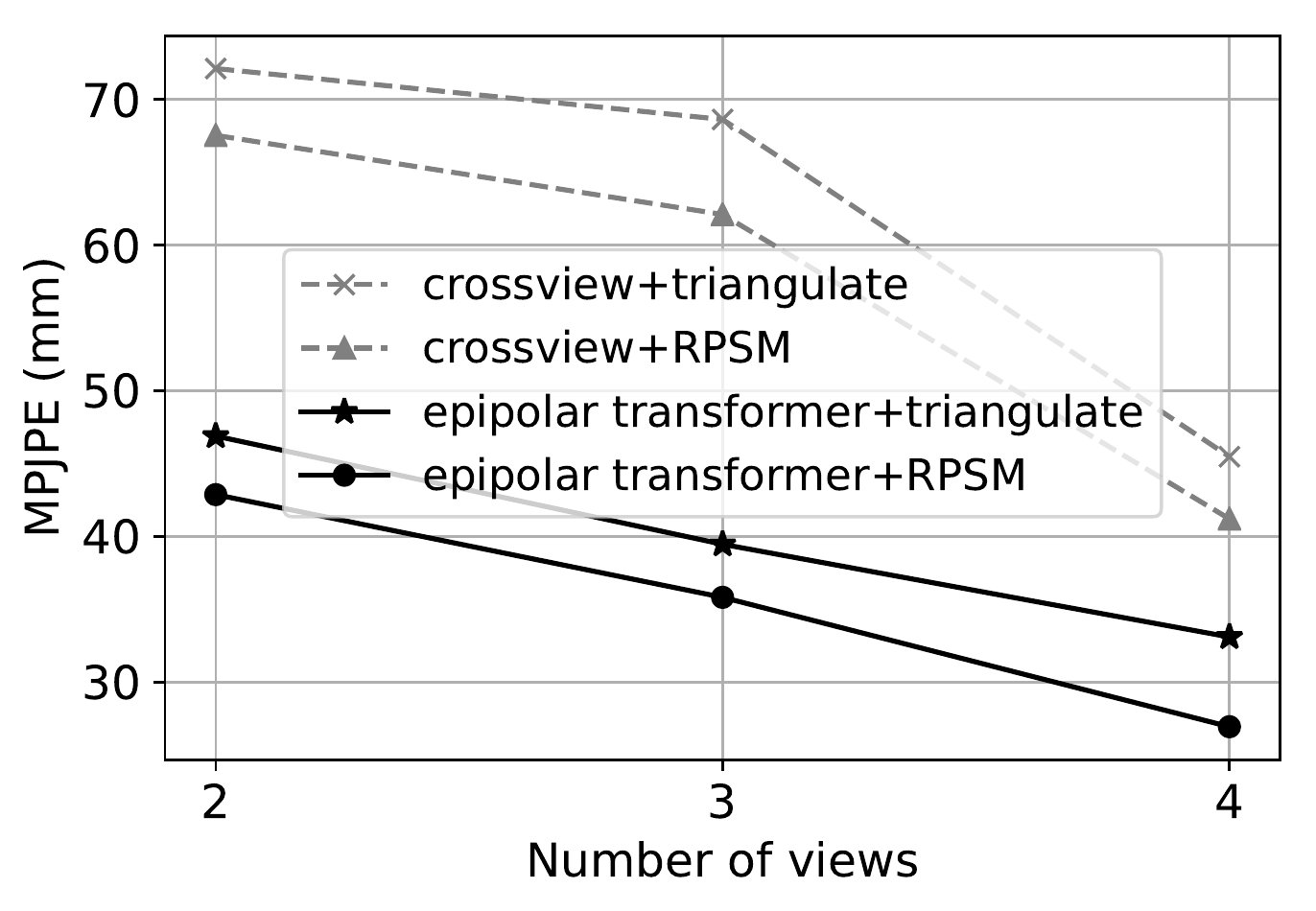}
    \caption{MPJPE by varying the number of views on \humandataset.}
    \label{fig:nviews_h36m}
% \end{center}
\end{figure}

\paragraph{Effect of the number of views used during test time:} As shown in \figref{fig:nviews_h36m}, compared with \crossview, the models with \ours\ still have better performance when there are fewer views. This shows that the \ours\ efficiently fuses features from other views. 

\paragraph{Comparison with \sota, no external datasets setting:} 
\tableref{tab:human_sota} shows the performance of several \sota\ methods when no external datasets are used. Our \ours\ outperforms the \sota\ by a large margin. Specifically, when using triangulation for estimating 3D human poses, \ours\ achieves \humansmallresnetfiftydlt\ mm, which is $\sim$ 12 mm better than \crossview\  while using the same backbone network (ResNet-50) and input size (256$\times$256). Using the recursive pictorial structural model (RPSM~\cite{qiu2019cross})  for estimating 3D poses, our \ours\ achieves \humansmallresnetfiftyrpsm\ mm, which is $\sim$ 14 mm better than the \crossview\  equivalent. Furthermore, adding \ours\ on ResNet-50 input size 256$\times$256 even surpasses the \sota\ result from \crossview\ on ResNet-152 input size 320$\times$320 by $\sim$ 4 mm, which is a 13\% relative improvement. 
Our model with data augmentation achieves MPJPE \humansmallresnetfiftydltaug\ mm with triangulation, which is better than the \sota\ without even requiring RPSM.
We believe one source of improvement comes from the fact that the \ours\ finds correspondences and fuses features dynamically based on feature similarity. This is more accurate than \crossview, which uses a static attention map for all input images from a pair of views.

% \begin{table}[]
% \centering
% \begin{tabular}{l|l}
% \hline
% method        & MPJPE           \\ \hline
% Multi-View Martinez~\cite{tome2018rethinking}& 57.0 \\
%     Pavlakos~\etal \cite{pavlakos2017harvesting} &   56.89\\
%     Tome~\etal~\cite{tome2018rethinking} & 52.8\\
%     Kadkhodamohammadi \& Padoy~\cite{kadkhodamohammadi2018generalizable} & 49.1\\ \hline
% R50\ \ \ 256$\times$256 + \crossview + triangulate   & 45.476          \\ \hline
% R50\ \ \ 256$\times$256 + crossview + \rpsm & 41.205          \\ \hline
% R152 320$\times$320 + \crossview+ triangulate &  38.29          \\ \hline
% R50\ \ \ 256$\times$256 + ours + triangulate & \textbf{\humansmallresnetfiftydlt} \\ \hline
% R152 320$\times$320 + crossview + \rpsm &  31.17          \\ \hline
% R50\ \ \ 256$\times$256 + ours + \rpsm & \textbf{\humansmallresnetfiftyrpsm} \\ \hline
% \end{tabular}
% \caption{Compare with ICCV'19 Crossview Fusion on \humandataset}
% \label{tab:human_sota}
% \end{table}

\begin{table*}
\setlength\tabcolsep{1.5pt}
  \resizebox{\textwidth}{!}{
   \begin{tabular}{lllllllllllllllll}
    \toprule
        MPJPE (mm)
    &  Dir &  Disc &  Eat &  Greet &  Phone &  Photo
    &  Pose &  Purch   &   Sit &  SitD &  Smoke &  Wait
    &  WalkD &  Walk &  WalkT &  \textbf{Avg}\\
     \midrule
    Multi-View Martinez~\cite{tome2018rethinking}  & 46.5 & 48.6 & 54.0 & 51.5 & 67.5 & 70.7 & 48.5 & 49.1  & 69.8 & 79.4 & 57.8 & 53.1 & 56.7 & 42.2 & 45.4 & 57.0 \\
    Pavlakos \etal \cite{pavlakos2017harvesting} &  41.2 &  49.2 &  42.8 &  43.4 &  55.6 &  46.9 &  40.3 & 
    63.7 & 97.6 &  119.0 &  52.1 &  42.7 &  51.9 &  41.8 &  39.4 &  56.9\\
    Tome \etal~\cite{tome2018rethinking} &  43.3 &  49.6 &  42.0 &  48.8 &  51.1 &  64.3 &  40.3 &  43.3 &  66.0 &  95.2 & 50.2 & 52.2 & 51.1 & 43.9 & 45.3 & 52.8\\
    Kadkhodamohammadi \& Padoy~\cite{kadkhodamohammadi2018generalizable}& 39.4&46.9&41.0&42.7&53.6&54.8&41.4&50.0& 59.9&78.8&49.8&46.2&51.1&40.5&41.0&49.1\\
    \hline
        R50\ \ \ 256$\times$256+triangulate & 38.9 &  46.1 &  36.2 &  59.7 &  46.4 &  44.7 &  44.9 &  37.7  &  51.2 &  72.0 &  48.2 &  61.0 &  46.2 &  45.7 &  52.0 & 48.7 \\
    R50\ \ \ 256$\times$256+crossview+triangulate\cite{qiu2019cross} &-&-&-&-&-&-&-&- &-&-&-&-&-&-&-& 45.5 \\
    R50\ \ \ 256$\times$256+ours+triangulate & 30.6 & 33.2 & 26.7 & 28.2 & 32.8 & 38.4 & 29.3 & 28.9  & 36.6 & 45.2 & 34.3 & 31.7 & 33.1 & 34.8 & 31.2 &  \humansmallresnetfiftydlt \\ 
\textbf{R50\ \ \ 256$\times$256+ours+triangulate \augmentation }  &29.0 &30.6 &27.4 &26.4 &31.0 &31.8 &26.4 &28.7   &34.2 &42.6 &32.4 &29.3 &27.0 &29.3 &25.9 &  \textbf{\humansmallresnetfiftydltaug} \\  \hline
R50\ \ \ 256$\times$256+crossview+\rpsm &-&-&-&-&-&-&-&-&-&-&-&-&-&-&-&41.2 \\ 

\textbf{R50\ \ \ 256$\times$256+ours+RPSM} & 25.7 & 27.7 & 23.7 & 24.8 & 26.9 & 31.4 &24.9 & 26.5 & 28.8 & 31.7 & 28.2 & 26.4 & 23.6 & 28.3 & 23.5 & \textbf{\humansmallresnetfiftyrpsm} \\ \hline
R152 320$\times$320+crossview+triangulate\cite{qiu2019cross} & 34.8 & 35.8 & 32.7 & 33.5 & 34.4 & 38.2 & 29.7 & 60.7 & 53.1 & 35.2 & 41.0 & 41.6 & 31.9 & 31.4 & 34.6 &  38.3 \\
R152 320$\times$320+crossview+RPSM &28.9 & 32.5& 26.6& 28.1&28.3&29.3&28.0&36.8 &42.0&30.5&35.6&30.0&29.3&30.0&30.5 &  31.2 \\
    % RANSAC (our implementation) & 24.1 & 26.1 & 24.0 & 24.6 & 27.0 & 25.0 & 23.3 & 26.8 & 31.4 & 49.5 & 27.8 & 25.4 & 24.0 & 27.4 & 24.1 & 27.4\\    
    % \textbf{Ours, algebraic (w/o conf)} & 22.9 & 25.3 & 23.7 & 23.0 & 29.2 & 25.1 & 21.0 & 26.2 & 34.1 & 41.9 & 29.2 & 23.3 & 22.3 & 26.6 & 23.3 & 26.9\\
    % \textbf{Ours, algebraic} & 20.4 & 22.6 & 20.5 & 19.7 & 22.1 & 20.6 & 19.5 & 23.0 & 25.8 & 33.0 & 23.0 & 21.6 & 20.7 & 23.7 & 21.3 & 22.6\\
    % \textbf{Ours, volumetric (softmax aggregation)} & \textbf{18.8} & \textbf{20.0} & 19.3 & 18.7 & \textbf{20.2} & \textbf{19.3} & 18.7 & 22.3 & 23.3 & 29.1 & \textbf{21.2} & \textbf{20.3} & \textbf{19.3} & \textbf{21.6} & \textbf{19.8} &\textbf{20.8}\\
    % \textbf{Ours, volumetric (sum aggregation)} &  19.3 & 20.5 & 20.1 & 19.3 & 20.6 & 19.8 & 19.0 & 22.9 & 23.5 & 29.8 & 22.0 & 21.4 & 19.8 & 22.1 & 20.3 & 21.3\\
    % \textbf{Ours, volumetric (conf aggregation)} &  19.9 & \textbf{20.0} & \textbf{18.9} & \textbf{18.5} & 20.5 & 19.4 & \textbf{18.4} & \textbf{22.1} & \textbf{22.5} &\textbf{ 28.7} & \textbf{21.2} & 20.8 & 19.7 & 22.1 & 20.2 & \textbf{20.8}\\
    \hline
    \bottomrule
   \end{tabular}
  }
 \caption[]{Comparison with \sota\ methods on \humandataset, where no additional training data is used unless specified. The metric is MPJPE (mm). "\augmentation": rotation and scaling augmentation. "-": models trained using released code~\cite{qiu2019cross}%\footnoteref{note:fuse}
 , where the per action MPJPE evaluation were not provided.
}
\label{tab:human_sota}
\end{table*}
% \footnotetext{https://github.com/microsoft/multiview-human-pose-estimation-pytorch/blob/master/INSTALL.md}

\paragraph{Comparison with state-of-the-art, with external datasets setting:}
%The results in \tableref{tab:human_sota} are trained only on \humandataset. 
\tableref{tab:human_extra} shows the performance of several \sota\ methods when external datasets are used.
Iskakov~\etal~\cite{iskakov2019learnable} established a 22.8 mm MPJPE RANSAC baseline with extra data from MS-COCO~\cite{coco} and MPII~\cite{andriluka20142d}. They further proposed the learnable weighted triangulation (algebraic w/ conf) and volumetric triangulation~\cite{NIPS2018_7755,moon2018v2v,Tung_2019_CVPR}, which achieve 19.2 mm and  17.7 mm respectively.
We fine-tune a MS-COCO+MPII pre-trained ResNet-152 384$\times$384 released in \cite{iskakov2019learnable} %\footnote{https://github.com/karfly/learnable-triangulation-pytorch} 
on \humandataset\ and achieve \humanextrabigresnetonefifydlt\ mm, which is the best among methods using vanilla triangulation. Besides, the \ours\ contributes very little to the number of parameters and computation. 
%Note that we did not fine-tune jointly on Human3.6M+MPII. You may notice that \crossview\ improves 5 mm and 10.4 mm respectively while fine-tuning using Human3.6M+MPII instead of Human3.6M. Fine-tuning with MPII help alleviate the issue of lack of diversity in the Human3.6M dataset. 
 
% Qiu~\etalcite{qiu2019cross} improves from  31.17 mm  MPJPE to 26.21 mm by joint training on \humandataset\ and MPII~\cite{andriluka20142d}.

\begin{table}[]
\setlength\tabcolsep{1pt}
\begin{tabular}{l|ll|cc|cc|l}
\hline
 & \multicolumn{2}{c|}{complexity} & \multicolumn{2}{c|}{pre-train} & \multicolumn{2}{c|}{fine-tune} & err. \\ \hline
 & param & MAC & COCO & MPII & H36M & MPII &  \\ \hline
crossview+tri. & 560M & 212B &  &  & \checkmark &  & 38.3 \\
crossview+RPSM & 560M & 212B &  &  & \checkmark &  & 31.2 \\
crossview+tri. & 560M & 212B &  &  & \checkmark & \checkmark & 27.9 \\
crossview+RPSM & 560M & 212B &  &  & \checkmark & \checkmark & 26.2 \\ \hline
triangulate & 69M & 204B & \checkmark & \checkmark & \checkmark & \checkmark & 22.8 \\
algebraic & 80M & 210B & \checkmark & \checkmark & \checkmark & \checkmark & 24.5 \\
algebraic w/ conf & 80M & 210B & \checkmark & \checkmark & \checkmark & \checkmark & 19.2 \\
volumetric$^+$ & 81M & 360B & \checkmark & \checkmark & \checkmark & \checkmark & \textbf{17.7} \\ \hline
ours+tri. & \textbf{69M} & \textbf{204B} & \checkmark & \checkmark & \checkmark &  & \humanextrabigresnetonefifydlt \\ \hline
\end{tabular}
\caption[]{Comparison with \sota\ methods using external datasets on \humandataset. +: data augmentation (\ie, cube rotation~\cite{iskakov2019learnable}). "err.": the error metric is MPJPE (mm). "tri." stands for triangulation. Number of Parameters and MAC (multiply-add operations) are calculated using THOP\footnotemark.}
\label{tab:human_extra}
\end{table}
\footnotetext{\href{https://github.com/Lyken17/pytorch-OpCounter}{github.com/Lyken17/pytorch-OpCounter}}

\subsection{Limitations}
The biggest limitation of our method is the reliance on precise geometric camera calibration. Poor calibration will lead to inaccurate epipolar lines thus incorrect feature matching.
Another limitation of our method is that the viewing angle of neighboring camera views should not be too large, otherwise there is a high likelihood a 3D point might be occluded in one of the views, which will make feature matching more difficult.

\section{Conclusion}

%2D poses is crucial for 3D keypoint estimation. 
We proposed the \ours, which enables 2D pose detectors to leverage 3D-aware features
through fusing features along the epipolar lines of neighboring views.
Experiments not only show improvement over the baseline on \humandataset\ and \handdataset, but also demonstrate that our method can improve multi-view pose estimation especially when there are few cameras.
Qualitative analysis of feature matching along the epipolar line also show that the \ours\ can provide more accurate matches in difficult scenarios with occlusions.
Finally, the \ours\ has very few learnable parameters and outputs features with the same dimension as the input, thus enabling it to be easily augmented to existing 2D pose estimation networks. 
For future work, we believe that the \ours\ can also benefit 3D vision tasks such as deep multi-view stereo \cite{yao2018mvsnet}.

\nocite{IonescuSminchisescu11}

{\small
\bibliographystyle{ieee_fullname}
\bibliography{egbib}
}

\clearpage
\appendix

\begin{figure*}
    \centering
\textbf{\Large Supplementary Materials: \OURS\ for Multi-view Pose Estimation}
\end{figure*}

\begin{figure}
\centering
    \includegraphics[width=\linewidth]{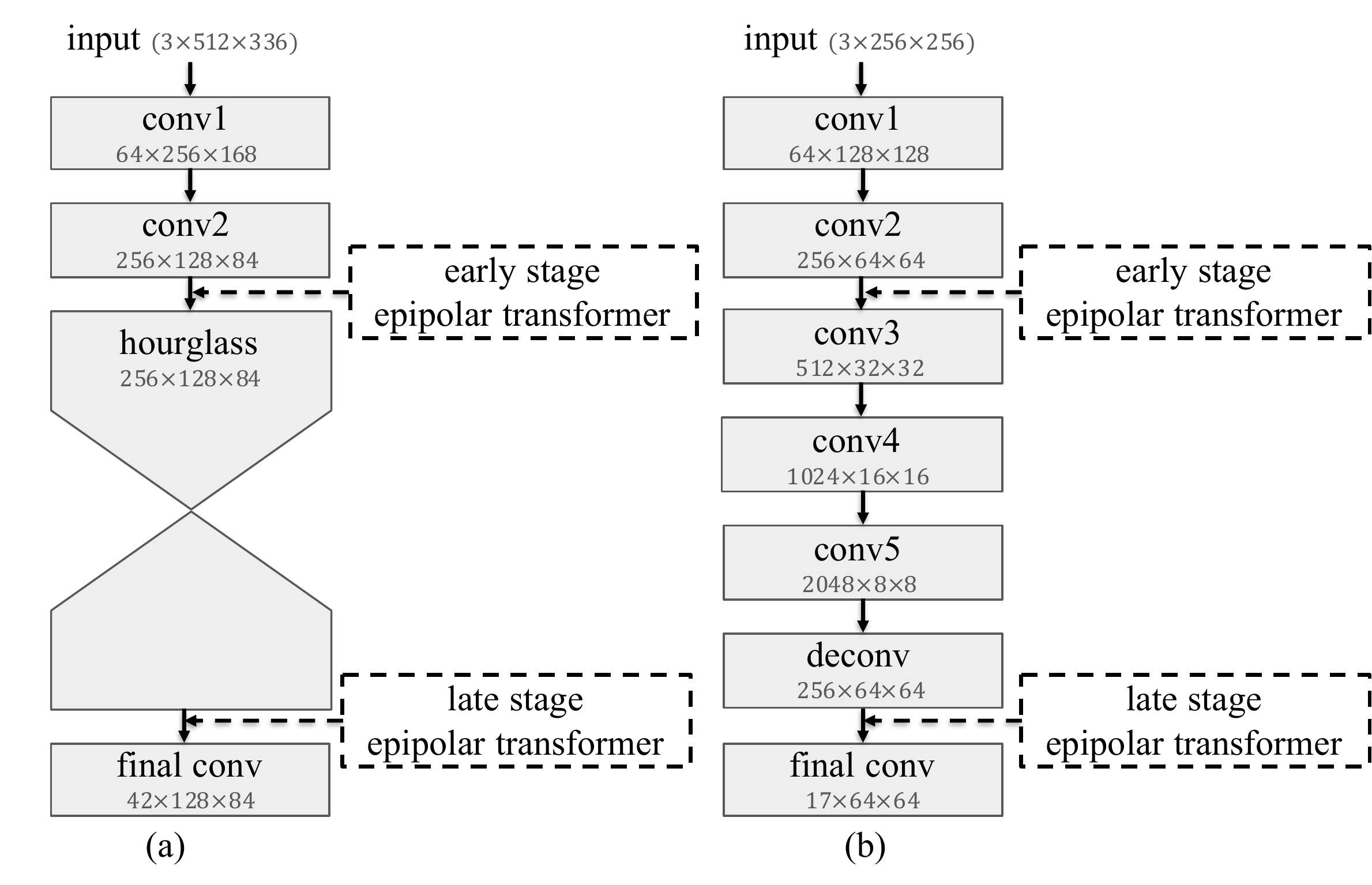}
    \caption{Early stage or late stage where we can add \ours\ to the backbone model. (a) Hourglass networks~\cite{newell2016stacked} on \handdataset . (b) ResNet-50 detector~\cite{xiao2018simple} on \humandataset .}
    \label{fig:hourglassresnet}
\end{figure}

\section{Dealing with Image Transformations}
\label{sec:augmentation}
As the \ours\ relies on camera calibration, any spatial transformation applied on the image needs 
to be reflected in the calibration parameters. 

\paragraph{Data augmentation:} Data augmentation like rotation, scaling and cropping can still be performed with the \ours. The projection matrix needs to be updated accordingly when the image is transformed with an affine transformation parameterized by $A \in \mathbb{R}^{2 \times 2}$ and $b \in \mathbb{R}^2$:
\begin{align}
M := \begin{bmatrix}
A & b \\
\mathbf{0}^T & 1 \\
\end{bmatrix}M
\end{align}
%scaled with $s_x$, $s_y$ and cropped with $t_x$, $t_y$
% \begin{align}
% M := \begin{bmatrix}
% s_x & 0 & t_x \\
%  0 & s_y & t_y \\
% 0 & 0 & 1 \\
% \end{bmatrix}M
% \end{align}
Different scaling and cropping parameters can be applied separately to the \mainview\ and \otherview. 

\paragraph{Scaling of projection matrices:} 
We should pay special attention to scale the projection matrices for resizing or pooling images.

Suppose the input image is spatially down-sampled $s_x$ and $s_y$ times along the x-axis and y-axis, the projection matrix is updated as follows:
\begin{align}
M := \begin{bmatrix}
1/s_x & 0 & (1 - s_x)/2s_x \\
 0 & 1/s_y & (1 - s_y)/2s_y \\
0 & 0 & 1 \\
\end{bmatrix}M
\end{align}
The coordinates are aligned with the center of pixels rather than the top-left corners, which is important for extracting features at precise locations in the \ours.

\section{2D Prediction Visualization in Video}
% In \href{https://drive.google.com/open?id=1UWiDrFutXIdmCeDX_zwqsf9s4M9n6l0L}{skeletons$\_$1min.mp4}, 
In \href{https://youtu.be/wY57gCBPP3U}{skeletons$\_$1min.mp4}, 
we visualize 2D predicted skeletons on \humandataset\ testing set. All methods are with ResNet-50~\cite{resnet} and image size 256$\times$256, corresponding to the following three entries in \tableref{tab:human_sota} respecitively: 
\begin{enumerate}
    \item R50\ \ \ 256$\times$256 + triangulate
    \item R50\ \ \ 256$\times$256 + crossview\cite{qiu2019cross} + triangulate
    \item R50\ \ \ 256$\times$256 + ours + triangulate    
\end{enumerate}

% In \href{https://drive.google.com/file/d/1dnUQD1I2qpo0fMdOqFHWBqFZqYjVcwRm/view}{twohand$\_$30.mp4}, 
In \href{https://youtu.be/151DlbnS3oY}{twohand$\_$30.mp4}, 
we visualize 2D predicted hands on \handdataset\ testing set. The video consists of the visualizations of the ground truth, the 5.46 mm baseline and 4.91 mm model with \ours\ in \tableref{tab:hand_inference}.

\section{Stages to Add \OURS}
In our experiments, we compared the performance between adding \ours\ in the early stage and adding in the late stage (see \tableref{tab:hand_stage}, paragraph \textbf{Which stage to insert the \ours}). \figref{fig:hourglassresnet} illustrates the precise places of inserting the \ours\ for the ``early'' and ``late'' settings in an one-stage Hourglass network~\cite{newell2016stacked} on \handdataset\ and ResNet-50~\cite{resnet} simple baseline~\cite{xiao2018simple} on \humandataset\ respectively.

% \begin{figure}
% \centering
% % \begin{center}
%     \includegraphics[width=0.8\linewidth]{imgs/hourglass.pdf}
%     \caption{Early stage or late stage where we can add \ours\ in hourglass network~\cite{newell2016stacked} on \handdataset}
%     \label{fig:hourglass}
% % \end{center}
% \end{figure}

% \begin{figure}
% \centering
% % \begin{center}
%     \includegraphics[width=0.8\linewidth]{imgs/resnet.pdf}
%     \caption{Early stage or late stage where we can add \ours\ in ResNet-50 simple baseline~\cite{xiao2018simple} on \humandataset}
%     \label{fig:resnet}
% % \end{center}
% \end{figure}

\section{\OURS\ Visualization on \humandataset}
\begin{figure*}
    \centering
    \includegraphics[width=0.7\linewidth]{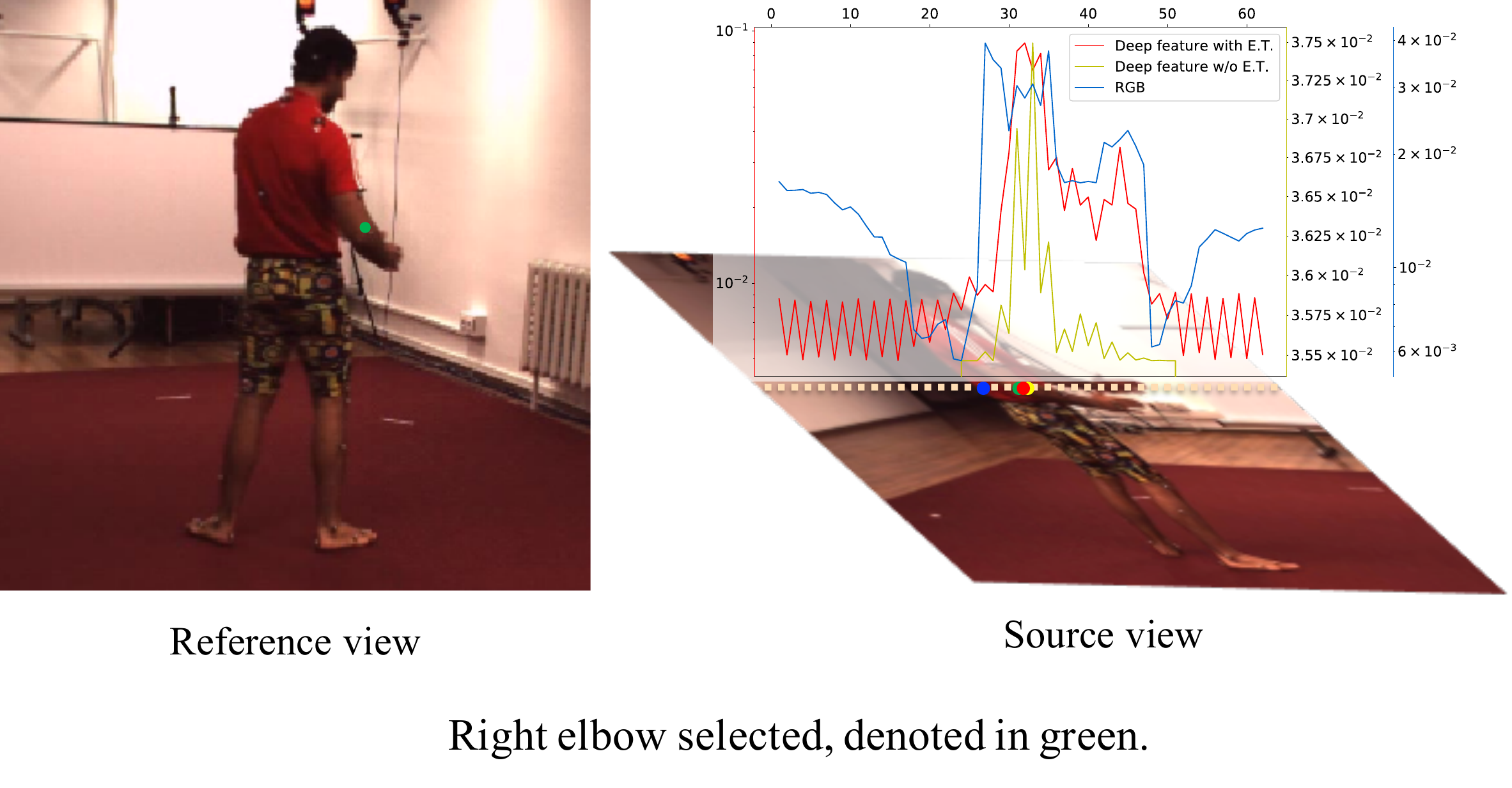}
    \caption{Visualizations of the matching results along the epipolar line in an \textbf{easy case} in \humandataset. We here use E.T. as a shorthand for \ours. The compared features are (a) deep features learned through the epipolar transformer (deep features with E.T., denoted in red), (b) deep feature learned by ResNet-50\cite{resnet} without epipolar transformer (deep features w/o E.T., denoted in yellow), and (c) RGB features (denoted in blue). Green dot on the reference view is the selected joint, and the green dot on the source view is the corresponding point offered by the groundtruth.}
    \label{fig:matching-comp-hm36-easy}
\end{figure*}

\begin{figure*}
    \centering
    \includegraphics[width=0.7\linewidth]{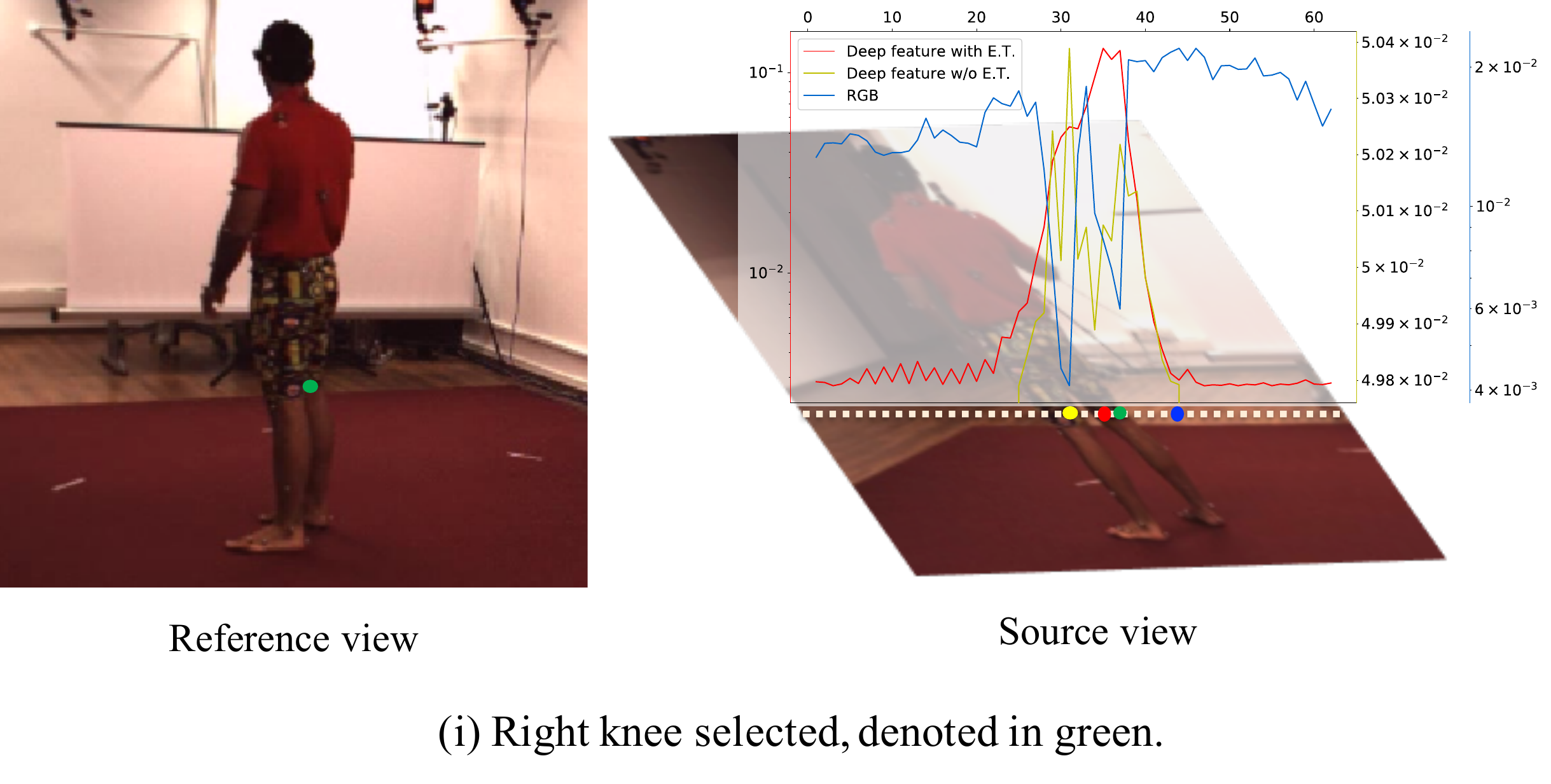}  
    \vspace{1cm}
    \caption{Visualizations of the matching results along the epipolar line in a \textbf{more difficult case} in \humandataset. We here use E.T. as a shorthand for Epipolar Transformer. The compared features are (a) deep features learned through the epipolar transformer (deep features with E.T., denoted in red), (b) deep feature learned by ResNet-50\cite{resnet} without epipolar transformer (deep features w/o E.T., denoted in yellow), and (c) RGB features (denoted in blue). Green dot on the reference view is the selected joint, and the green dot on the source view is the corresponding point offered by the groundtruth.}
    \label{fig:matching-comp-hm36}
\end{figure*}

\begin{figure*}
    \centering
    \includegraphics[width=0.7\linewidth]{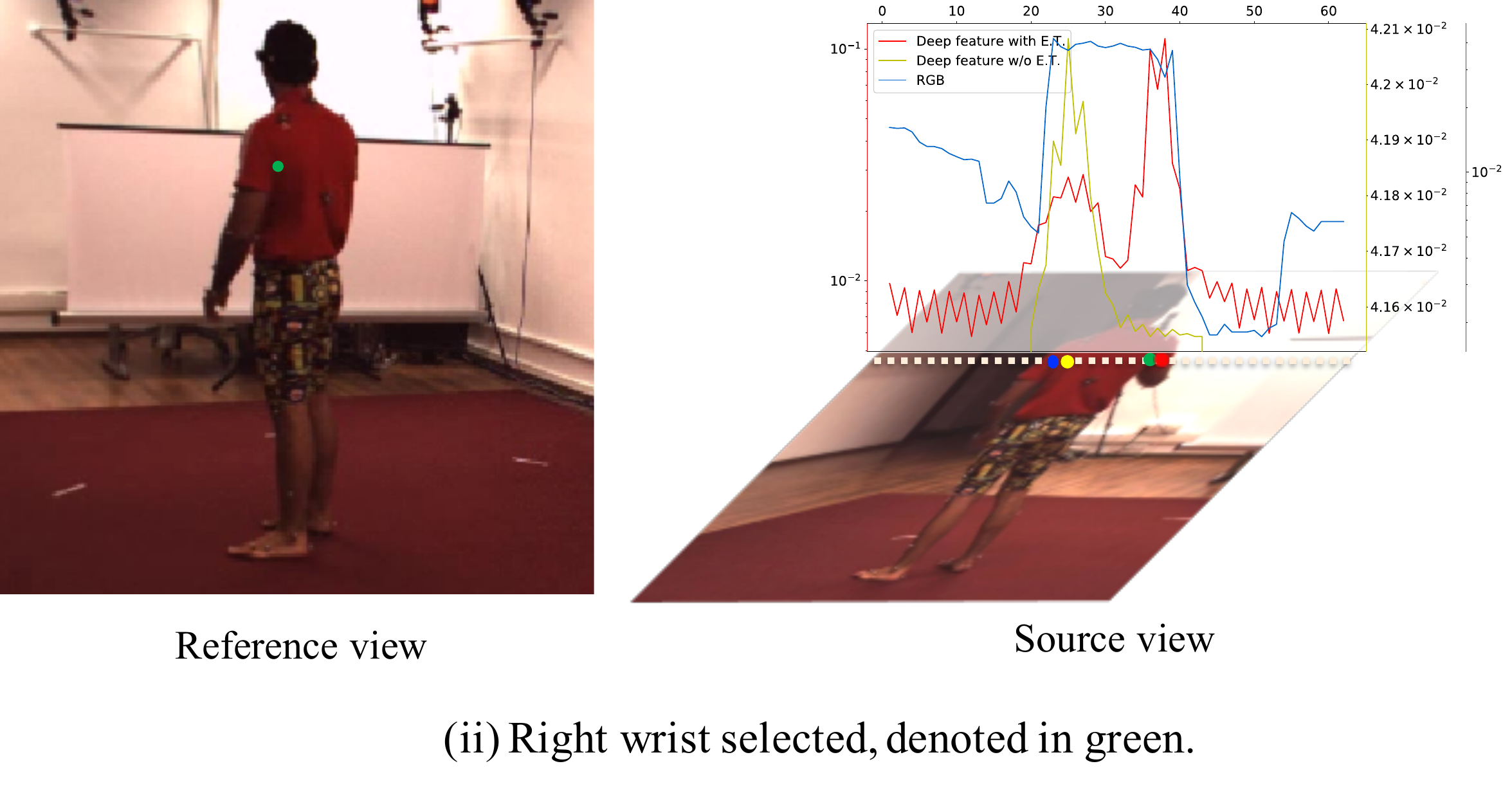}  

    \includegraphics[width=0.7\linewidth]{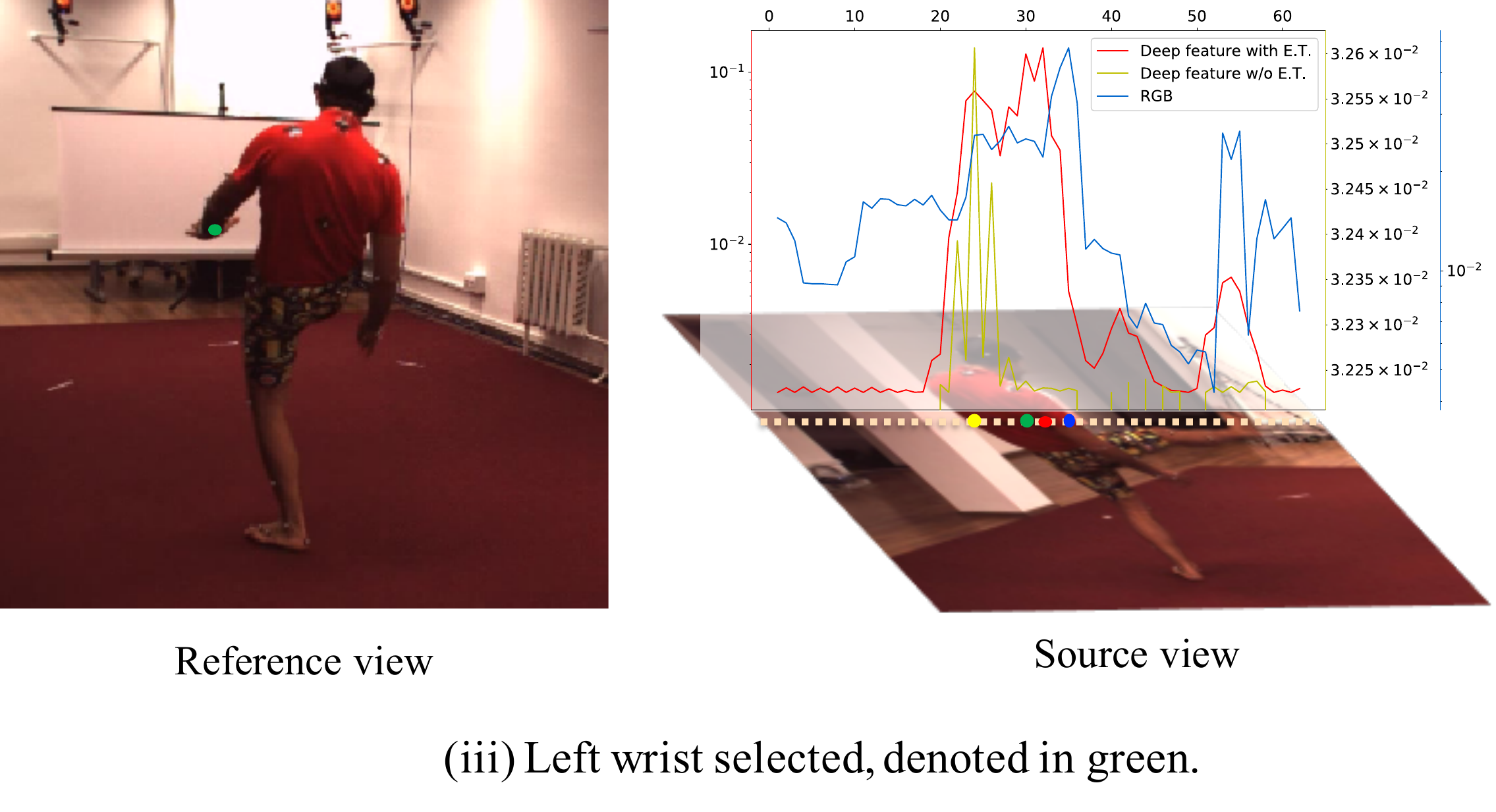}  
\caption{(Cont')Visualizations of the matching results along the epipolar line in \textbf{more difficult cases} in \humandataset. We here use E.T. as a shorthand for Epipolar Transformer. The compared features are (a) deep features learned through the epipolar transformer (deep features with E.T., denoted in red), (b) deep feature learned by ResNet-50\cite{resnet} without epipolar transformer (deep features w/o E.T., denoted in yellow), and (c) RGB features (denoted in blue). Green dot on the reference view is the selected joint, and the green dot on the source view is the corresponding point offered by the groundtruth.}
    \label{fig:matching-comp-hm36-2}
\end{figure*}

We show the visualizations of feature matching similarity for images from \humandataset\ in \figref{fig:matching-comp-hm36-easy}, \figref{fig:matching-comp-hm36}, and \figref{fig:matching-comp-hm36-2}.  In \figref{fig:matching-comp-hm36-easy}, we show a visualization of the matching results of an easy case in \humandataset. Note that in this case, our prediction aligns well with the ground truth. Using the features from the baseline, that is, deep features extracted without using the \ours, the matched point denoted by the yellow dot is also accurate.
However, for more difficult cases in \figref{fig:matching-comp-hm36} and \figref{fig:matching-comp-hm36-2}, the deep features learned with the \ours\ can perform more accurate matching.
%As shown in \figref{fig:matching-comp-hm36-2}, we also tested the matching on more difficult cases with occlusions. Our predictions with \ours\ (red dot) are closer to the ground truth points compared to the features without the awareness of the multi-view information.

% {\color{citecolor} Green}
% {\color{Dandelion} yellow}
% {\color{blue} blue}
% {\color{red} red}

\section{\OURS\ Visualization on \handdataset}
\begin{figure*}
\centering
    % \begin{center}
    \includegraphics[width=0.4\linewidth]{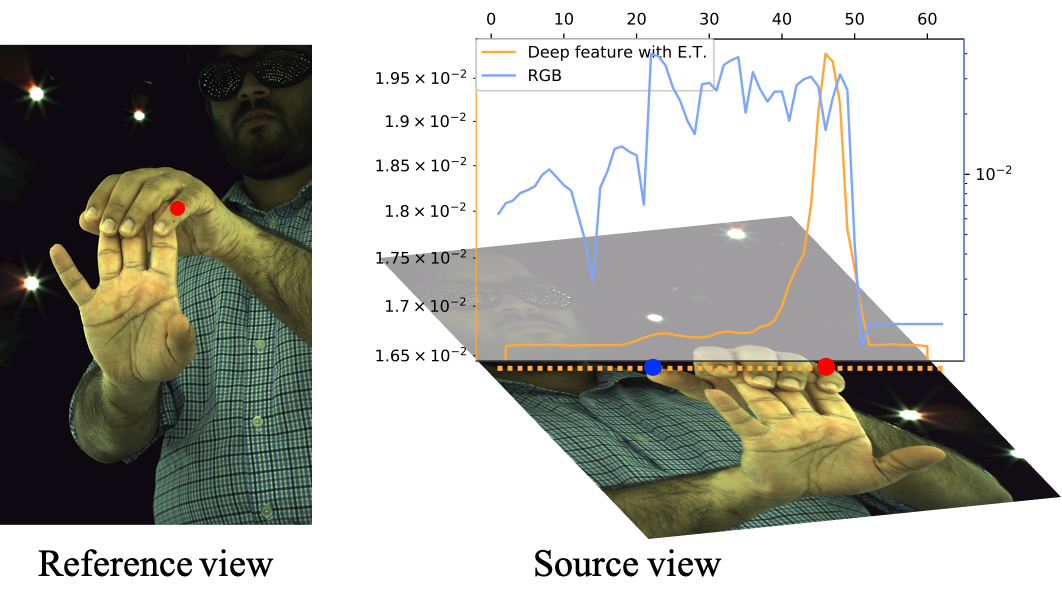}
    \hspace{1cm}
    \includegraphics[width=0.4\linewidth]{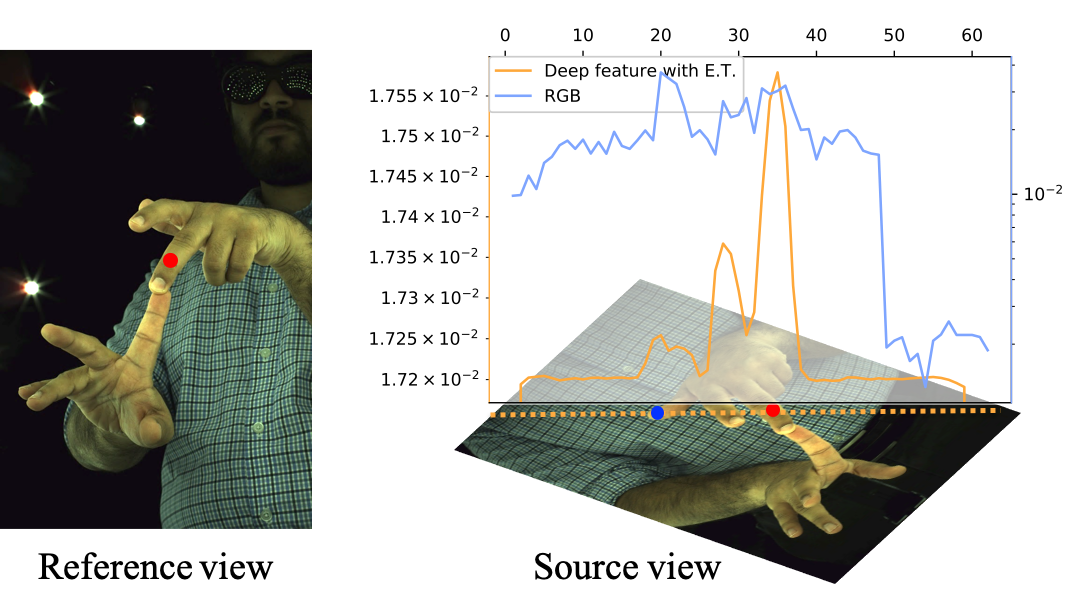}
    \caption{Comparison of feature-matching results along the epipolar line. The compared features are color features and the deep features learned through the \ours\ (deep features with E.T.). The best matches of the \ours\ (red) and RGB (blue) are shown on the epipolar lines. The similarity distributions along the epipolar lines are also shown. %(\textit{best viewed in color}) 
    }
    \label{fig:visualization}
    % \end{center}
\end{figure*}

We show visualization of feature matching similarity for images from \handdataset.
As shown in \figref{fig:visualization}, the color features have multiple peaks in similarity due to the different fingers having similar colors.
On the other hand, for most cases deep features trained through the \ours\ are able to discriminate the correct finger from all the other similar looking fingers.

\end{document}